\documentclass[transmag]{IEEEtran}
\usepackage{latexsym}
\usepackage{graphicx}
\usepackage{amsfonts,amssymb,amsmath}
\usepackage{makecell}
\usepackage{tabularx}
\usepackage{hyperref}
\usepackage{algorithm}
\usepackage{algorithmic}
\usepackage{color}
\usepackage{bm}
\def\BibTeX{{\rm B\kern-.05em{\sc i\kern-.025em b}\kern-.08em T\kern-.1667em\lower.7ex\hbox{E}\kern-.125emX}}
\begin{document}
\title{End-to-end Compression Towards Machine Vision: Network Architecture Design and Optimization}

\author{Shurun Wang, Zhao Wang, Shiqi Wang, Yan Ye
\thanks{This research was supported in part by the National Natural Science Foundation of China under 62022002, in part by the Hong Kong Research Grants Council, Early Career Scheme (RGC ECS) under Grant 21211018, General Research Fund (GRF) under Grant 11203220.}
\thanks{S. Wang and S. Wang are with the Department of Computer Science, City University of Hong Kong, Hong Kong, China (e-mail: srwang3-c@my.cityu.edu.hk;  shiqwang@cityu.edu.hk).}
\thanks{Z. Wang and Y. Ye are with the Alibaba Group (e-mail: baixiu.wz@alibaba-inc.com;  yan.ye@alibaba-inc.com).}

}

\iffalse
\IEEEmembership{Member, IEEE}
\thanks{This paragraph of the first footnote will contain the 
date on which you submitted your paper for review. It will also contain 
support information, including sponsor and financial support acknowledgment. 
For example, ``This work was supported in part by the U.S. Department of 
Commerce under Grant BS123456''.}
\thanks{The next few paragraphs should contain 
the authors' current affiliations, including current address and e-mail. For 
example, F. A. Author is with the National Institute of Standards and 
Technology, Boulder, CO 80305 USA (e-mail: author@ boulder.nist.gov).}
\thanks{S. B. Author, Jr., was with Rice University, Houston, TX 77005 USA. He is 
now with the Department of Physics, Colorado State University, Fort Collins, 
CO 80523 USA (e-mail: author@lamar.colostate.edu).}
\thanks{T. C. Author is with 
the Electrical Engineering Department, University of Colorado, Boulder, CO 
80309 USA, on leave from the National Research Institute for Metals, 
Tsukuba, Japan (e-mail: author@nrim.go.jp).}
\fi

\IEEEtitleabstractindextext{\begin{abstract}
%The compression of visual data has been a long-standing research topic and inspiring progress has been achieved recently, especially fueled by deep learning. In spite of compression performance improvement achieved, existing end-to-end compression algorithms are still dedicatedly designed, targeting for better signal fidelity in terms of rate-distortion optimization. In this paper, we further improve the compression performance towards machine vision with the exploration of network architecture and the joint optimization with generalized rate-accuracy distortion. More specifically, we explore the proposed inverted bottleneck structure for end-to-end compression towards machine vision, which could efficiently extract and represent the semantic information for analysis. Moreover, an end-to-end joint optimization framework is proposed, which optimizes the compression and analysis models semoustanously  and the representation capability is further improved by embedding the analytic accuracy into the optimization scheme. Object detection plays the fundamental role in various intelligent analysis applications and is selected as a showcase of end-to-end compression towards machine vision. Extensive experiments show that significant BD-rate savings could be achieved in terms of analysis performance with less computational complexity and we also demonstrate the promise of the proposed scheme with strong generalization capability for various tasks, due to the enabling of signal-level reconstruction.

The visual signal compression is a long-standing problem. Fueled by the recent advances of deep learning, exciting progress has been made. Despite better compression performance, existing end-to-end compression algorithms are still designed towards better signal quality in terms of rate-distortion optimization. In this paper, we show that the design and optimization of network architecture could be further improved for compression towards machine vision. We propose an inverted bottleneck structure for the encoder of the end-to-end compression towards machine vision, which specifically accounts for efficient representation of the semantic information. Moreover, we quest the capability of optimization by incorporating the analytics accuracy into the optimization process, and the optimality is further explored with generalized rate-accuracy optimization in an iterative manner.
%and a principled framework is further proposed to explore the optimality with generalized rate-accuracy optimization in an iterative manner. 
We use object detection as a showcase for end-to-end compression towards machine vision, and extensive experiments show that the proposed scheme achieves significant BD-rate savings in terms of analysis performance. Moreover, the promise of the scheme is also demonstrated with strong generalization capability towards other machine vision tasks, due to the enabling of signal-level reconstruction. 
\end{abstract}


}

\maketitle

\section{INTRODUCTION}

\IEEEPARstart{R}{ecent} years have witnessed an explosive growth of visual data  
%in the era of artificial intelligence (AI) on account of the 
driven by the unprecedented proliferation of multimedia acquisition, processing and display devices. It has been reported that visual data accounts for the largest proportion in the global data traffic \cite{cisco2019cisco}. As such, the compact representation of visual data is highly demanded and has been extensively studied towards human perception based on a series of human visual system (HVS) characteristics. With the notable success of deep learning in various visual analysis and understanding tasks, machine has become an alternative and increasingly important terminal for ultimate consumption of visual data. This is grounded on the widely accepted view that traditional pipelines where manpower is relied to process and analyze the huge data volume are impractical for real-time applications such as smart cities and intelligent security. According to the prediction of Cisco, machine-to-machine (M2M) communications will occupy half of the internet connections between global devices \cite{cisco2020cisco}.  

The utilization of the visual data towards machine vision primarily relies on how the data could be efficiently represented in a compact way. Facing such arising challenges of video compression for machine (VCM), numerous methods have been developed beyond the traditional video coding schemes. The prominent paradigm is referred as Analyze-then-Compress (ATC) \cite{redondi2013compress}, which is developed based on the fact that features could be more compact than signals. This is in stark contrast with the traditional Compress-then-Analyse (CTA) paradigm, as the features are compressed in ATC whereas the visual signals are compressed in CTA. 
Since machine vision relies on features for understanding and analysis, compact feature representation can dramatically reduce the visual data representation expense and facilitate various intelligent applications with high throughput between front-end devices such as sensors and back-end systems such as central servers. More specifically, the standards of Compact Descriptors for Visual Search (CDVS) \cite{duan2015overview} and Compact Descriptors for Video Analysis (CDVA) \cite{duan2018compact} standardized highly compact descriptors for images and videos. The exploration of video coding for machine (VCM) has also been launched by Moving Picture Experts Group (MPEG), in an effort to extend the compact representation to various artificial intelligent tasks towards machine vision \cite{xia2020emerging}. 

Clearly, the ATC paradigm which dramatically economizes the representation cost does not ensure the reconstruction at the signal level and significantly affects the generalization capability for various tasks. This has been an important aspect ignored by ATC, which lacks strong generalization capability across different analytics tasks. More importantly, human involved monitoring, a step that is still indispensable when the event is of sufficient interest, cannot be supported. As such, the ATC and CTA work in their own way towards compact visual information representation, and an advanced scheme that fulfills both purposes is highly desirable. 
Herein, we propose a customized scheme for visual signal compression scheme towards machine vision grounded on the end-to-end deep image coding framework. The proposed scheme is designed and optimized to fully accommodate for the characteristics of machine vision, and moreover enjoys the advantages of both high generalization capability and better rate-accuracy performance. Overall, the main contributions of the paper are as follows, 
\begin{itemize}
\item We propose an inverted bottleneck structure from the perspective of channel number distribution for the encoder of the end-to-end coding towards machine vision. The proposed structure is motivated by the fact that the semantic high-level features are more important in compact representations, and leads to lower encoding complexity without degrading the rate-accuracy performance. 
\item We propose an encoder optimization scheme to improve the rate-accuracy performance. The potentials of end-to-end compression for machine vision are extensively explored with the proposed generalized rate-accuracy optimization framework. 
%We propose an encoder optimization scheme to improve the rate-accuracy performance. The optimality is achieved in a scientifically sound way, such that the potentials of end-to-end compression for machine vision have been fully explored. 
\item We carry out extensive experiments based upon the object detection task to evaluate the performance in terms of rate-accuracy, showing superior performance of the proposed scheme.  Meanwhile, the superiority of the proposed scheme over traditional ATC and CTA approaches is also verified through this process. %and computational complexity
\end{itemize}

The rest of the paper is organized as follows. In Section 2, we review and summarize the related works. In Section 3, we introduce the whole compression and analysis pipeline based upon the end-to-end compression framework. 
The proposed architecture customized for machine vision is introduced in Section 4, including the motivations, design philosophy and principles. In Section 5, we present the optimization towards machine vision with off-line search. The discussions {regarding the connections} of the proposed method with image compression, feature compression and visual analysis are provided in Section 6. 
In Section 7, the experimental results are presented to show the efficiency of proposed schemes in terms of rate-accuracy and the encoding simplification in terms of the number of parameters and encoding time. We conclude the paper in Section 8.

\section{Related Works}
\begin{figure*}[htbp]
\centerline{\includegraphics[width=7.2in]{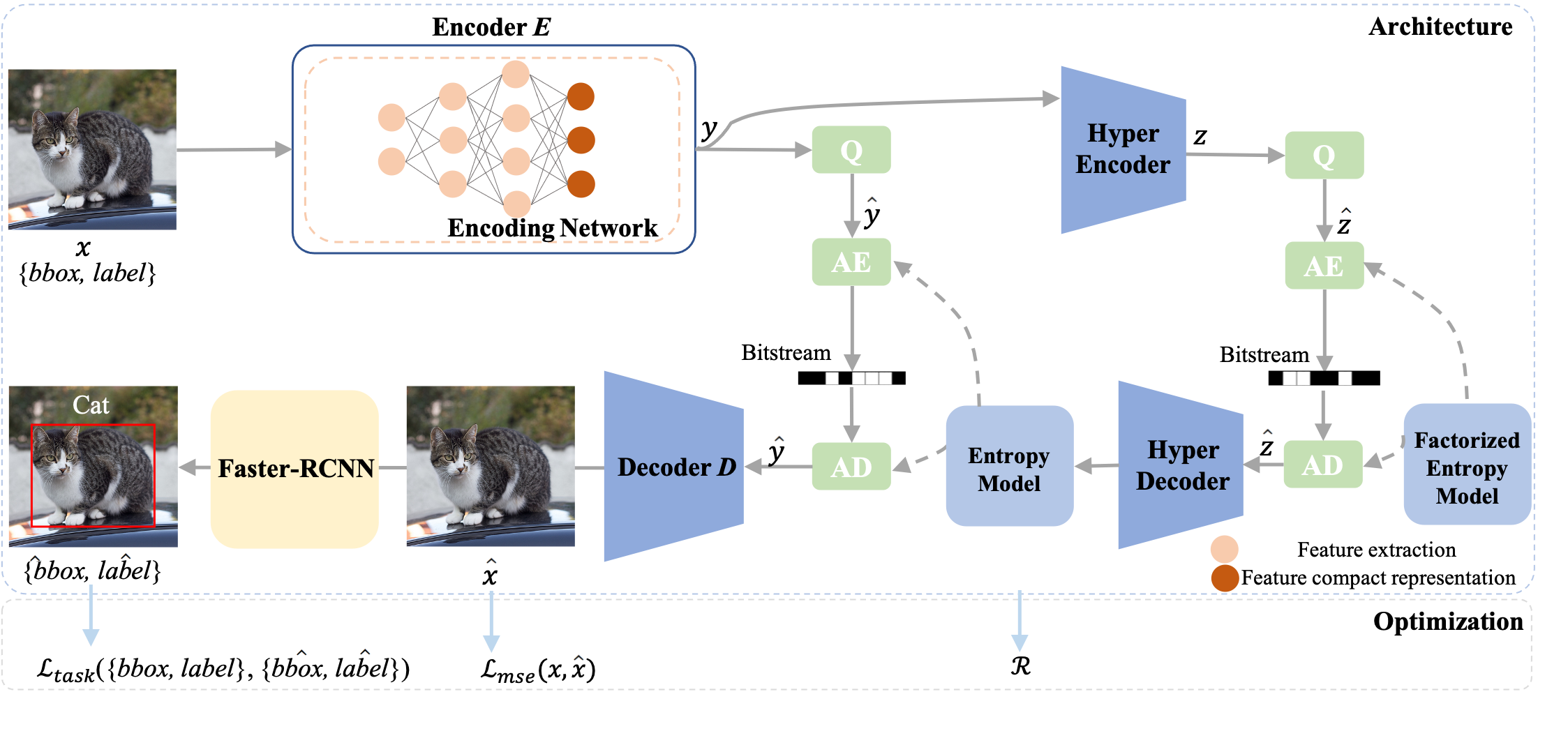}}
\vspace{-2mm}
\caption{The pipeline of the end-to-end image compression towards machine vision. More specifically, object detection is selected as the show case of the machine vision task. The encoder network is specifically designed, and the distortion of signal reconstruction $\mathcal{L}_{mse}$, the object detection loss towards machine vision $\mathcal{L}_{task}$ and the representation expense $\mathcal R$ can be acquired for joint optimization.}
\label{architecture}
\vspace{-4mm}
\end{figure*}
How the visual signals can be compactly represented towards different objectives has been a long-standing problem. Central to such problem is the maximization of the ultimate utility within available bit rate, where the utility can be defined in different manners depending on the final receiver of the visual signals. 
Recently, various methods have been proposed, which can be mainly classified into visual signal compression and compact feature representation. 
\subsection{Visual Signal Compression}%The compression of visual data towards HVS}
The traditional visual signal coding is primarily characterized based on the throughput of the channel and perceived distortion
of the reconstructed signals. 
In particular, the image/video coding techniques are driven by the development of visual data compression standards. 
More specifically, JPEG \cite{wallace1992jpeg}, JPEG 2000 \cite{rabbani2002jpeg2000} and WebP \cite{lian2012webp} have been developed to encode the still images, and H.264/AVC \cite{wiegand2003overview}, H.265/HEVC \cite{sullivan2012overview} and the state-of-the-art standards including Versatile Video Coding (VVC) \cite{choi2019design} and AVS3 \cite{zhang2019recent} are developed for video compression. Central to the image and video coding is the rate-distortion optimization (RDO), which is pursued to optimize the coding efficiency with operational control of the encoder \cite{ramchandran1994rate, sullivan1998rate, stankowski2015rate,karczewicz2008rate}. In the research of video coding, in addition to deriving the best Lagrangian multiplier~\cite{li2008laplace}, the distortion measure has also been replaced towards better modeling of the visual perception~\cite{wang2011ssim}. 

The recent advancements of deep learning have achieved substantial breakthroughs in visual computing tasks. Benefiting from the strong representation capability of neural networks, deep learning has also been applied to image compression, achieving rapid progress in coding performance. The recurrent neural network (RNN) based image compression was first proposed, achieving comparable performance with JPEG \cite{toderici2015variable}. Motivated by the discrete cosine transform (DCT) in traditional image compression, a block based deep learning transformation was proposed, which outperforms JPEG at low bit rates \cite{liu2018cnn}. Inspired by the local gain control, the generalized divisive normalization (GDN) was proposed in \cite{balle2015density}, and an end-to-end image compression based upon GDN was developed, surpassing the performance of JPEG 2000 \cite{balle2016end}. The statistical dependency is further exploited with a variational hyper-prior model \cite{balle2018variational}. Based on this method, a joint autoregressive and hierarchical prior for image compression was proposed in \cite{minnen2018joint}, achieving performance improvement comparing with the state-of-the-art image codecs. Although prominent improvement has been achieved for visual signal representation, the optimization in the compression process is still based upon HVS instead of machine vision. As such, there are still 
many challenges regarding efficient and compact representation of visual signals towards machine vision in the era of artificial intelligence.

\subsection{Compact Visual Feature Representation}
The ATC paradigm was developed grounded on the
widely rooted view that visual analytics and understanding primarily rely on features instead of textures. Moreover, the features are much more compact than textures, enabling the simultaneous transmission of the videos to the central cloud and redistribution of the computational load between front-end and back-end. 
To facilitate the visual retrieval in real-world applications, the standards of Compact Descriptors for Visual Search (CDVS) \cite{duan2015overview} and Compact Descriptors for Video Analysis (CDVA) \cite{duan2018compact} have been completed. 
Based on a series of deep neural networks as the backbone for the deep learning feature extraction, such as AlexNet \cite{krizhevsky2012imagenet}, VGG \cite{simonyan2014very}, Inception Net \cite{szegedy2015going} and ResNet \cite{he2016deep}, various algorithms have been proposed. More specifically, the philosophy of hybrid video coding has been introduced to the joint compression model for the local and global deep learning features in \cite{ding2020joint}. An end-to-end deep learning feature compression with multi-granularity constraint and teacher-student learning was proposed in \cite{wang2020end}, achieving performance improvement in terms of rate-analysis accuracy. A lossy intermediate deep learning feature compression towards intelligent sensing was proposed in \cite{chen2020toward}, which provides a prominent paradigm for the standardization of deep learning feature compression. Moreover, efforts have been devoted to performing image understanding tasks directly on the compact representations for the DNN-based compression methods \cite{torfason2018towards}.
The exploration of video coding for machine (VCM) \cite{duan2020video, vcmcfe} has also been launched by MPEG to plot a whole picture for the compact representation of visual signals towards machine vision.

\section{The Pipeline of Compression towards Machine Vision}
For machine vision, the high-level semantic information extracted from the network plays a critical role. This is in stark contrast with traditional compression which targets the visual signal reconstruction. 
The proposed scheme, which is built upon the end-to-end compression framework, preserves the advantage of the signal level reconstruction for high generalization capability, and is customized for machine vision. 
%We first introduce the pipeline of compression and analytics, and subsequently propose the encoding network architecture, including the motivations, design philosophy and merits. 
Herein, the object detection is adopted as the show case of the machine vision task as it plays a fundamental role in various artificial intelligent applications such as intelligent transportation, smart city and intelligent industry. This also aligns with the use cases and technical requirements for VCM, since object detection has been included into the required properties of algorithms under various tasks, as indicated in \cite{vcmuse}.
%The corresponding proposal \cite{vcmuse} has also been approved by the working group of MPEG.
Moreover, object detection plays fundamental roles in the high-level understanding of visual signal, such as event detection \cite{ke2007event}, anomaly detection \cite{basharat2008learning} and tracking \cite{balaji2017survey}.
%For image compression, the encoder is responsible for feature extraction and compression. As such, in order to design a network structure towards machine vision, we combine the machine vision task with end-to-end image compression. Moreover, we conduct a comprehensive study of the encoder structure for proposed end-to-end image compression and further propose an inverted bottleneck structure towards machine vision. 
%\subsection{}

The whole pipeline of the end-to-end image compression towards machine vision is shown in Fig. \ref{architecture}. The architecture and optimization both play indispensable roles in compression. The architecture is composed of an end-to-end image compression codec and a model for the machine vision task. For image compression, the latent representation $y$ of the original image $x$ is the output of encoder $E$, which is subsequently quantized with $Q$ as $\hat{y}$. The decoded image $\hat{x}$ is reconstructed with decoder $D$. In order to capture spatial dependencies in the latent representation $y$, the hyper-latent $z$ is acquired with hyper encoder and utilized by hyper decoder after quantization. Moreover, the bitrate $\mathcal{R}$ is estimated by the Shannon entropy and the distribution could be modeled with the entropy and factorized entropy model without context model, as proposed in \cite{minnen2018joint}.
%of the distribution $p_{\hat{y}}$, which is modeled by the entropy model.
For machine vision task, the object detection model Faster-RCNN \cite{ren2015faster} is adopted and the results of object detection, the predicted bounding boxes and labels $\{\hat{bbox}, \hat{label}\}$, could be obtained given the decoded image $\hat{x}$. Herein, our scheme is built upon the typical end-to-end coding framework in \cite{minnen2018joint}. Within this framework, the encoder network architecture and optimization method are specifically designed for the machine vision tasks and introduced in the subsequent sections. 

\section{Encoding Network Architecture}
\begin{figure}[bt]
\centerline{\includegraphics[width=3.4in]{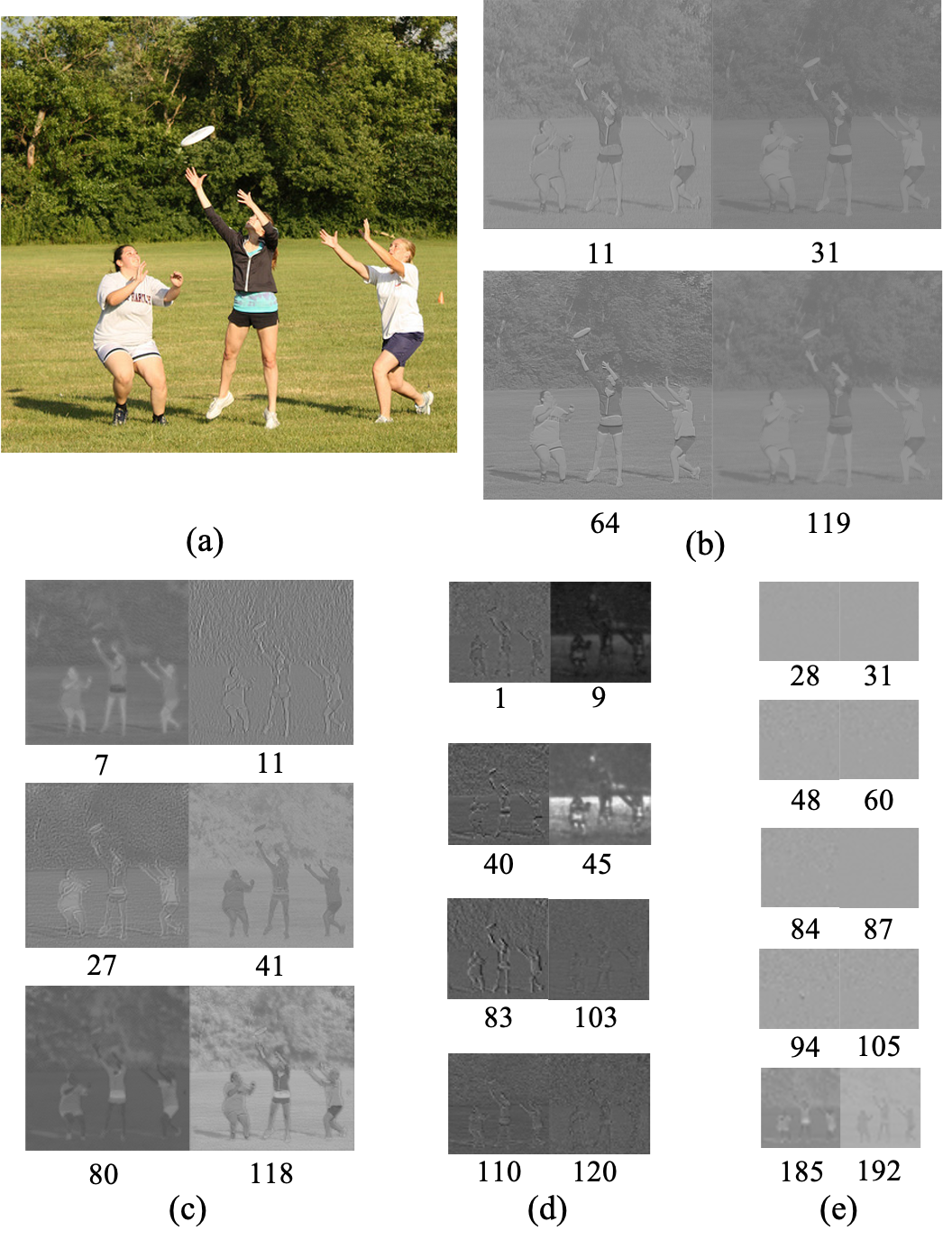}}
\vspace{-2mm}
\caption{(a) The original image. (b)$\sim$(e) Visualizations of the feature maps of various convolutional layers of a typical end-to-end encoder \cite{minnen2018joint} by means of min-max normalization. For each layer, the channels are randomly selected.}
\label{feature}
\vspace{-4mm}
\end{figure}

\begin{figure}[bt]
\centerline{\includegraphics[width=3.3in]{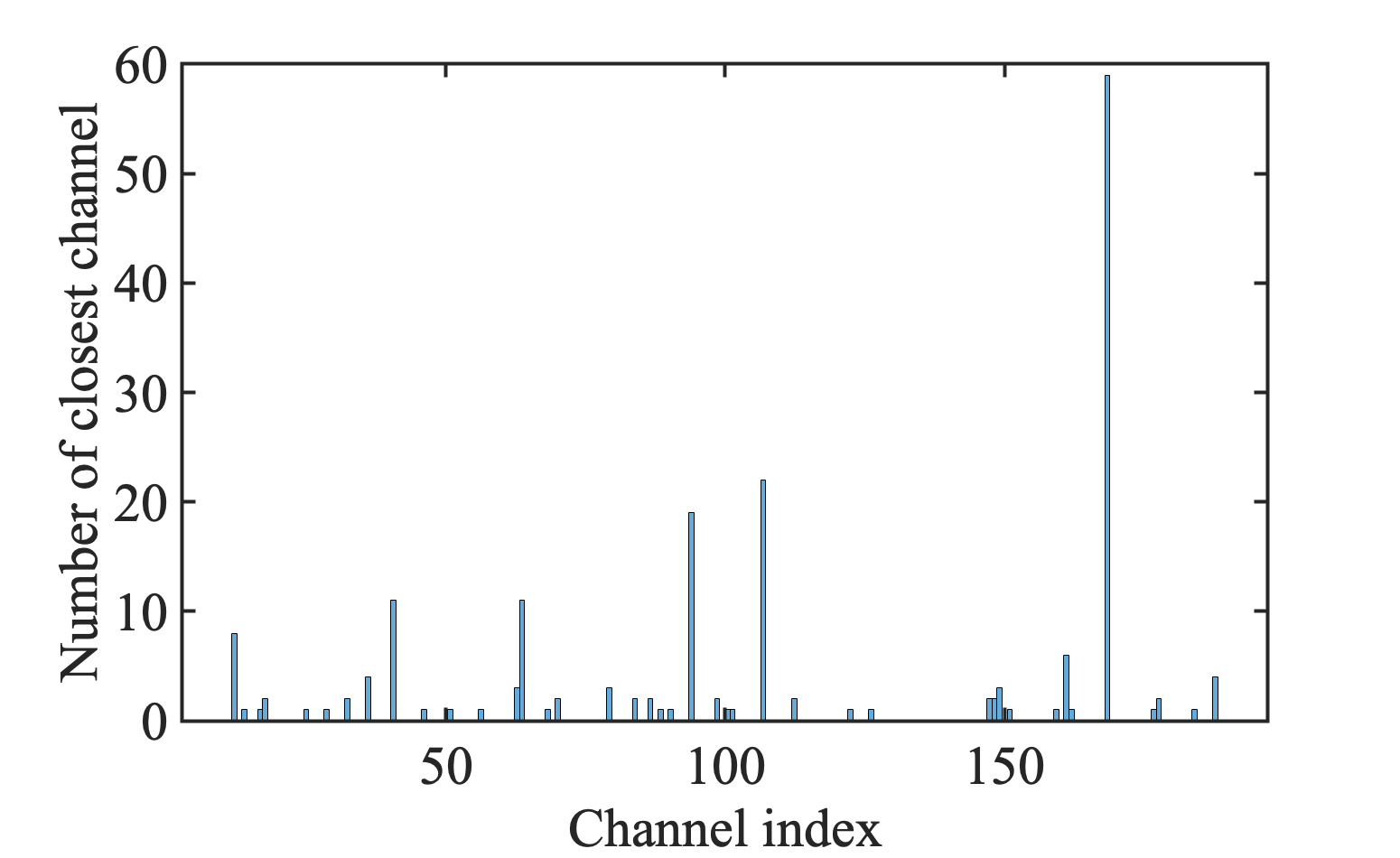}}
\vspace{-2mm}
\caption{Illustration of the channel redundancy of the $4^{th}$ convolutional layer for a typical end-to-end encoder \cite{minnen2018joint}. %in terms of minimal MAD. 
Specifically, the horizontal axis indicates the channel index ranging from 1 to 192. The vertical axis represents the number of closest channels for every channel index. In particular, the closest channel is selected by calculating the MAD between the current channel and every other channel, and finally the channel with the minimal MAD is selected. The range of MAD is from 0.0010 to 0.1646.}
\label{conv4}
\vspace{-4mm}
\end{figure}

\begin{figure}[bt]
\centerline{\includegraphics[width=3.5in]{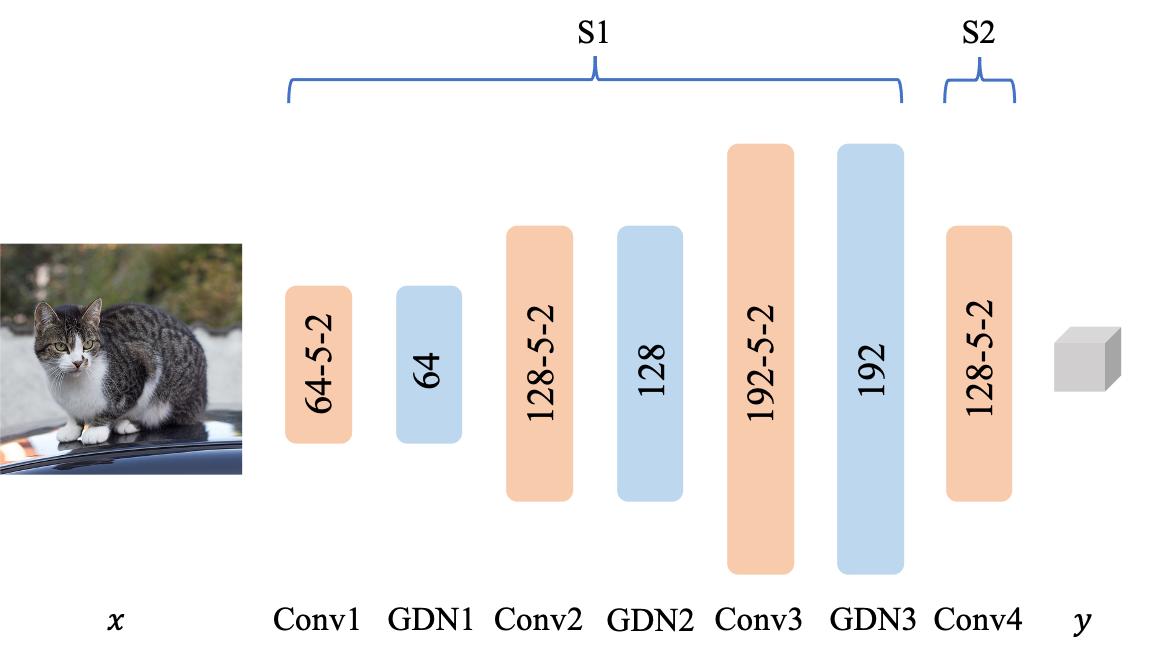}}
\vspace{-2mm}
\caption{The inverted bottleneck structure of the end-to-end image compression encoder towards machine vision. ``Conv'' and ``GDN'' represent the convolutional layers and the GDN layers respectively. The $N-K-S$ annotations indicate the channel number $N$, the kernel size $K$ and the stride $S$ of the convolutional layer respectively.}
\label{encoder}
\vspace{-4mm}
\end{figure}

In principle, the encoder is responsible for generating the latent code that accounts for the compact representation of the original images. The optimization objective of the end-to-end codecs has always been the quality towards human vision perception. However, semantic information is more important for compression towards machine vision. Such inconsistency could result in representation redundancy and computational inefficiency for the compression towards machine vision, especially for the front-end devices such as cameras and mobilephones, where the encoder is deployed with limited computation power  \cite{chen2020toward}. In particular, there are four convolutional layers in the encoder of the end-to-end image compression. As shown in Fig. \ref{feature}, we visualize the feature maps in various layers {of encoder in \cite{minnen2018joint}} with min-max normalization. {There is an obvious delamination phenomenon among the first three layers. Moreover, the similarity among the channels of the last layer is also investigated with the mean absolute difference (MAD). Specifically, for every channel $c_{i}$, the channel with the minimal MAD is 
selected, denoted as $c_{j}, j\neq i$.} As shown in Fig. \ref{conv4}, it is interesting to find that around 59 channels share the same most similar channel, revealing abundant redundancy.
%There is obvious redundancy in the feature maps in multiple layers, the repeated structural information in the first several layers and the abundant near-plain feature maps in the last layer. In order to tackle this problem
Motivated by this, we propose an inverted bottleneck structure for the encoder towards machine vision to achieve a compact representation with high efficiency and low complexity, as shown in Fig. \ref{encoder}. Specifically, the layers of the encoder have been divided into two stages: feature extraction and compact representation{, denoted as S1 and S2 respectively}. The channel number increases in the feature extraction stage to extract semantic features with high diversity {and low complexity} for machine vision task, and the channel number decreases in the compression stage to achieve a compact representation, serving as the output of encoder.

As illustrated in \cite{zeiler2014visualizing, he2016deep}, the delamination representation in deep learning models enables the deep layers to extract the high-level semantic information that is very abstract for facilitating the understanding. By contrast, texture information mainly exists in the first two convolutional layers with high redundancy, as shown in Fig. \ref{feature}~(b)\&(c). The features in the third convolutional layer mainly reveal the discriminative information, such as edges and structures, which are crucial for machine vision task, as shown in Fig. \ref{feature} (d). 
Such visualization and analysis further support the design philosophy of the inverted bottleneck structure in terms of the channel number in the encoder, which decouples the feature representation and compact representation to some extent, and emphasizes the semantic information for analysis. There are several advantages of the proposed architecture. First, the proposed inverted bottleneck structure reduces the computational redundancy. 
For the first stage of feature extraction, the channel numbers of the first two layers are reduced to economize computational cost, and the channel number of the third layer is retained to preserve the semantic information for machine vision. Second, the proposed inverted bottleneck structure can also eliminate the redundancy of compact representation. The second stage of encoder targets at the compact representation of extracted semantic features from the first stage. As shown in Fig. \ref{feature}(e), only several feature maps contain discriminative information representation. As such, the channel number of the last layer decreases to achieve an efficient representation towards machine vision.

\section{Generalized Rate-Accuracy Optimization}
Generally speaking, image compression is primarily characterized based upon the coding bits and distortion, and the fundamental issue is to obtain the best trade-off between them. Herein, the typical rate-distortion optimization (RDO) problem is converted to rate-accuracy optimization (RAO), based on the fact that the ultimate receiver is the machine vision. Such RAO problem has been widely studied in feature compression and joint texture-feature compression~\cite{li2018joint, zhang2016joint, ma2018joint, ding2017rate}. In this work, we attempt to take a further step to develop the generalized RAO which holds promise to improve both generalization capability and feasibility.  

\subsection{Generalized RAO Framework}
The RAO framework, which serves as the foundation of the loss function, is formulated as follows,
\begin{equation}
\begin{aligned}
\mathcal{L}&=\lambda_{1}\mathcal{D}+\mathcal{R}, 
\end{aligned}
\label{equ1}
\end{equation}
where \(\mathcal{D}\) denotes the generalized distortion which involves both signal fidelity and analyses accuracy, and \(\mathcal{R}\) denotes the coding bits. The Lagrange multiplier \(\lambda_1\) controls the tradeoff between
\(\mathcal{R}\) and \(\mathcal{D}\), and we follow the typical end-to-end coding for setting the Lagrange multiplier \cite{begaint2020compressai}. 
%Regarding that machine is the ultimate consumer of visual signal, semantic information is much more important than the reconstruction quality at pixel wise. As such, in an effort to propose an end-to-end image compression towards machine vision, 
Central to this optimization problem is the definition of \(\mathcal{D}\), as it is generally acknowledged that the final 
quality can only be as good as what it is optimized for. As the final receiver is assumed to be the machine vision, the 
optimization goal should be the analysis accuracy. However, such optimization requires the deterministic network model and parameters for analysis, while there is a lack of the generic deep learning model that could be applied to a broad range of analysis tasks. This motivates us to incorporate the signal level distortion into \(D\), which is given by, 
\begin{equation}
\begin{aligned}
\mathcal{D}&=\mathcal{L}_{mse}+\lambda_{2}\mathcal{L}_{task},
\end{aligned}
\end{equation}
where the task loss $\mathcal{L}_{task}$ indicates the accuracy in machine vision, which is typically defined as \cite{ren2015faster} in object detection. 
Herein, we introduce the mean squared error (MSE) between the original and reconstructed images $\mathcal{L}_{mse}$, which is incorporated as the regularization term for modeling the signal-level distortion. This indicates that the signal quality is simultaneously preserved, which improves the generalization capability of the proposed framework. The parameter $\lambda_{2}$ is the weighting factor that controls the signal level distortion and accuracy.

\subsection{Generalized Distortion Modeling}
In this subsection, we introduce the searching strategy to obtain the most appropriate \(\lambda_2\) in practice.  
It is worth mentioning that there are infinite candidates of $\lambda_{2}$ values, while it is impractical to perform the global search for the optimal one. By contrast, the 
empirical selection of $\lambda_{2}$ can also limit the performance. To tackle this problem, we propose the  optimization based on the off-line searching, in an effort to achieve enhanced performance towards machine vision. 
%rate-distortion optimization, which has been widely applied to conventional visual signal compression, can be applied to efficient candidate search and achieve better performance improvement. In an effort to achieve the optimality towards machine, we propose an off-line search in terms of RDO. 

%\textbf{The algorithm of off-line search} 
%For every quality level, it determines a weighting distribution between distortion and coding expense under a specific compression quality level, 
Herein, we assume that there are \(n\) quality levels in total, referred to as $q_{i}, 1\leq i \leq n$. Various values of $\lambda_{1}$ lead to multiple quality degradation levels in compression, which are denoted as the quality level. For each quality level, it corresponds to a set of potential \(\lambda_2\) values, and the optimal one is selected based on the rate-accuracy (RA) cost. To reduce the number of potential \(\lambda_2\) values in RA cost calculation in order to reduce the amount of tedious training and testing, we propose the iterative approach. In particular, 
%point of every quality level $q_{i}$ is selected empirically, denoted as $\lambda_{2,i_{0}}$ with an interval $d, d\geq0$. 
%Subsequently, 
the candidate set of \(\lambda_2\) for quality $q_{i}$ at the $t^{th}$ iteration is determined with a central point $\lambda_{2,i}$ and an interval $d$, denoted as $\bm{\lambda_{2,i_{t}}}=\{\lambda_{2,i_{t}}-d,~\lambda_{2,i_{t}},~\lambda_{2,i_{t}}+d\}=\{\lambda_{2,i_{t},1},~\lambda_{2,i_{t},2},~\lambda_{2,i_{t},3}\}$. The initial values of $\lambda_{2,t}$ and $d$ are empirically selected.
%where  $\lambda_{2,i_{0}}$ is selected empirically and \(d\) is the interval. 
For $t^{th}$ iteration and quality level $q_{i}$, the end-to-end image compression is optimized with every candidate in $\bm{\lambda_{2,i_{t}}}$ and the object detection is performed on the decoded images for the RA cost calculation. Herein, in RA cost calculation, we adopt the evaluation metric in MPEG VCM \cite{vcmtest}, where 
the mean average precision (mAP) with the intersection of union from 0.5 to 0.95 with interval 0.05 is selected as the performance metric of object detection, denoted as mAP@0.5:0.95 and referred as $map$ for convenience. As such, the accuracy variation of the object detection could be defined as,
\begin{equation}
    \mathcal{D}_{task}=(map(x)-map(\hat{x}))/map(x).
\end{equation}
where \(x\) denotes the original image and \(\hat{x}\) denotes the decoded image.

%In $t^{th}$ iteration, there are $3\times n$ distortion points in total, where distortion points are denoted as $\{(bpp, \mathcal{D}_{task})\}$. 
Herein, the RA cost $J_{i_{t},j}$ is defined as 
\begin{equation}
\begin{aligned}
    J_{i_{t},j}&=\gamma_{i_{t},j}\mathcal{R}_{i_{t},j}+\mathcal{D}_{task, i_{t},j}
\end{aligned}
\label{racost}
\end{equation}
where $i_{t},j$ denotes the $i^{th}$ quality level, $t^{th}$ iteration and the $j^{th}$ $\lambda_{2}$ candidate, $j=1,2,3$.
%, and $j=1,2,3$. 
Again, \(\gamma_{i_{t},j}\) denotes the Lagrangian multiplier and is obtained with the curve fitting with the cubic polynomial function of all RA cost points in the first iteration. For every iteration, we select the one with the minimal RA cost. 

A bidirectional search for the new candidates of $\lambda_{2}$ with interval $d$ is performed afterwards. Specifically, in $t^{th}$ iteration, the selected $\lambda_{2}$ with minimal RA cost for $i^{th}$ quality level is denoted as $\lambda_{2,i_{t}}^{*}$. If $\lambda_{2,i_{t}}^{*}=\lambda_{2,i_{t-1}}^{*}$, the interval $d$ should be shrunk with factor $w$, $d=d/w$. Otherwise, $d$ remains unchanged. The new candidates for the $t+1^{th}$ iteration at $i^{th}$ quality level could be $\bm{\lambda_{2,i_{t+1}}}=\{\lambda_{2,i_{t}}^{*}-d, \lambda_{2,i_{t}}^{*}, \lambda_{2,i_{t}}^{*}+d\}$. 
Such strategy ensures that 
the proposed algorithm could achieve a monotonically decreasing RA cost across iterations, finally terminate after finite procedures. 
The pipeline of the proposed algorithm is shown in Algorithm \ref{alg}.

\begin{algorithm}[bt]
%\textsl{}\setstretch{1.8}
	\renewcommand{\algorithmicrequire}{\textbf{Input:}}
	\renewcommand{\algorithmicensure}{\textbf{Output:}}
	\caption{The algorithm pipeline of the optimization based on the off-line searching.}
	\label{alg} %$\{\lambda_{2,i_{1},j}\}$
	\begin{algorithmic}
        \REQUIRE The start points of various quality levels, $\bm{\lambda_{2,i_{1}}}=\{\lambda_{2,i_{1},j}\}$, $1\leq i \leq n$ and $j=1,2,3$. The initial interval $d$, the shrink factor $w$, the maximal iteration number $N$ and $t=1$.
        \ENSURE The optimized $\lambda_{2}$ value at various quality levels, $\{\lambda_{2,i}\}$, $1\leq i \leq n$.
        \REPEAT 
        \STATE Optimize the  compression model with the $\lambda_{2}$ candidates in $t^{th}$ iteration at $i^{th}$ quality level, $\{\lambda_{2,i_{t},j}\}$,$j=1,2,3$.
        \STATE Evaluate the RA cost, defined in Equ. \ref{racost}, with the trained network, $\{(\mathcal{R}_{i_{t},j}, \mathcal{D}_{task,i_{t},j})\}$. 
        \STATE Select the $\lambda_{2}$ candidate with minimal RA cost, denoted as $\lambda^{*}_{2,i_{t}}$.
        \IF{$\lambda^{*}_{2,i_{t}}$=$\lambda^{*}_{2,i_{t-1}}$}
        \STATE $d=d/w$
        \ENDIF 
        \STATE The $\lambda_{2}$ candidates in $t+1^{th}$ iteration is $\bm{\lambda_{2,i_{t+1}}}=\{\lambda_{2,i_{t}}^{*}-d, \lambda_{2,i_{t}}^{*}, \lambda_{2,i_{t}}^{*}+d\}$.
        \STATE $t=t+1$.
        \UNTIL $t=N$
	\end{algorithmic}
%\vspace{-4mm}
\end{algorithm}

\begin{table}[tb]
\caption{The channel distributions at two stages. ``Cons'', ``Down'' and ``Up'' represent the constant, monotonically decreasing and monotonically increasing tendency at each stage respectively. The channel distributions of S2-C and S1-I are identical.}
\vspace{-2mm}
\setlength\tabcolsep{4pt}
\begin{tabular}{|p{0.9cm}<{\centering}|p{0.9cm}<{\centering}|p{0.9cm}<{\centering}|p{0.9cm}<{\centering}|p{0.9cm}<{\centering}|p{0.9cm}<{\centering}|p{0.9cm}<{\centering}|}
\hline
\multicolumn{1}{|c|}{} & \multicolumn{3}{c|}{Stage1 (Conv1,2,3)}    & \multicolumn{3}{c|}{Stage2 (Conv4)}      \\ \hline
Tendency               & Cons & Down & Up & Cons   & Down & Up \\ \hline
Notation               & S1-C     & S1-D     & S1-I     & S2-C & S2-D     & S2-I     \\ \hline
Conv1                  & 192      & 320      & 64       & 64         & 64       & 64       \\ \hline
Conv2                  & 192      & 256      & 128      & 128        & 128      & 128      \\ \hline
Conv3                  & 192      & 192      & 192      & 192        & 192      & 192      \\ \hline
Conv4                  & 192      & 192      & 192      & 192        & 128      & 256      \\ \hline
\end{tabular}
\label{invbt}
\vspace{-4mm}
\end{table}

\begin{figure*}[htbp]
\begin{minipage}[b]{0.3\linewidth}
  \centering
  \centerline{\includegraphics[width=6.0cm]{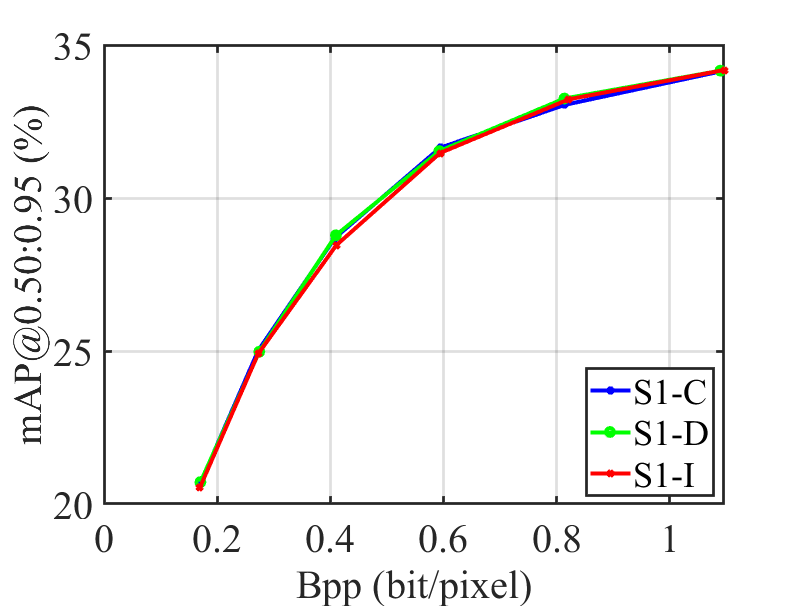}}
%  \vspace{1.5cm}
  \centerline{(a)}\medskip
\end{minipage}
\hfill
\begin{minipage}[b]{0.3\linewidth}
  \centering
  \centerline{\includegraphics[width=6.0cm]{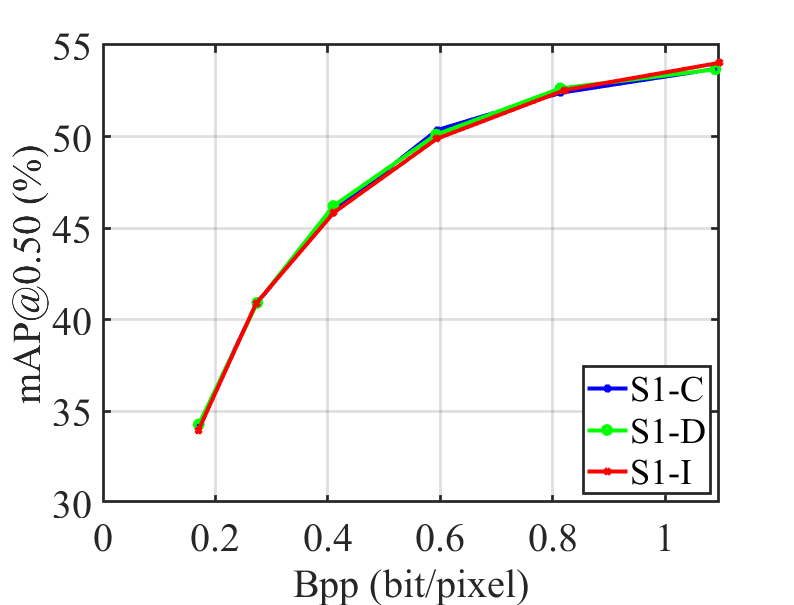}}
%  \vspace{1.5cm}
  \centerline{(b)}\medskip
\end{minipage}
\hfill
\begin{minipage}[b]{0.3\linewidth}
  \centering
  \centerline{\includegraphics[width=6.0cm]{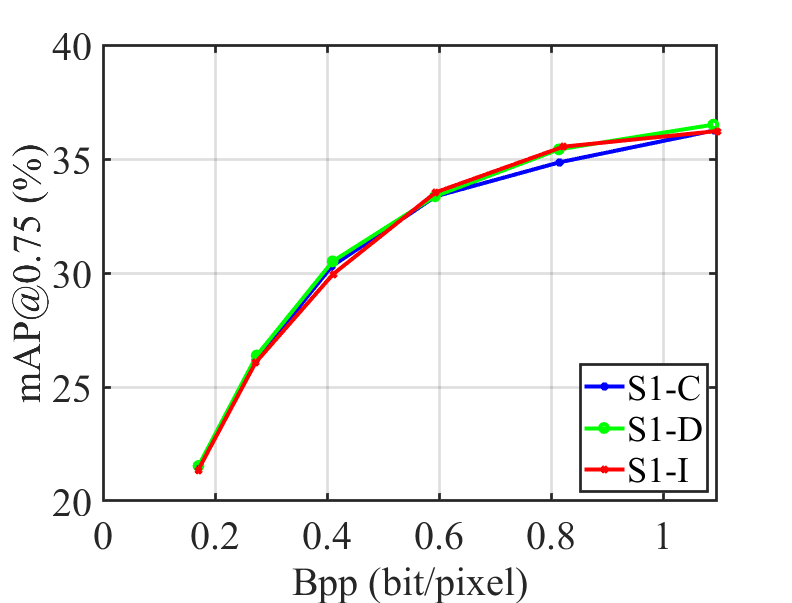}}
%  \vspace{1.5cm}
  \centerline{(c)}\medskip
\end{minipage}
\vspace{-4mm}
\caption{The performance comparison between the three distributions for stage 1 in terms of (a) rate-mAP@0.50:0.95; (b) rate-mAP@0.50; (c) rate-mAP@0.75.}
\vspace{-2mm}
\label{stage1}
\end{figure*}

\begin{figure*}[t]
\begin{minipage}[b]{0.3\linewidth}
  \centering
  \centerline{\includegraphics[width=6.0cm]{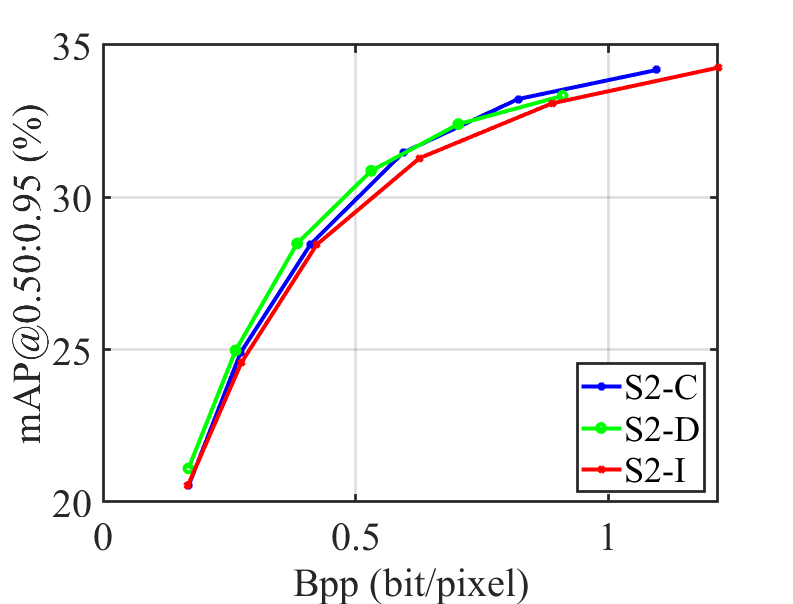}}
%  \vspace{1.5cm}
  \centerline{(a)}\medskip
\end{minipage}
\hfill
\begin{minipage}[b]{0.3\linewidth}
  \centering
  \centerline{\includegraphics[width=6.0cm]{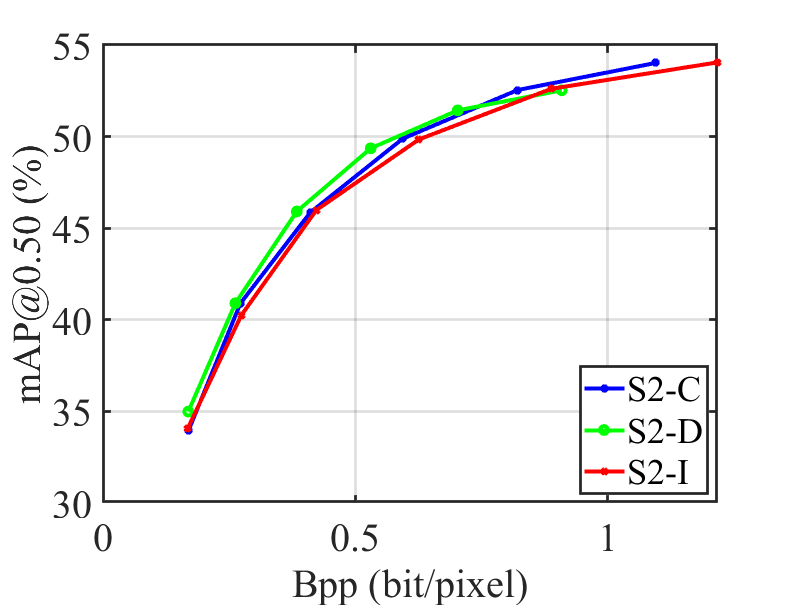}}
%  \vspace{1.5cm}
  \centerline{(b)}\medskip
\end{minipage}
\hfill
\begin{minipage}[b]{0.3\linewidth}
  \centering
  \centerline{\includegraphics[width=6.0cm]{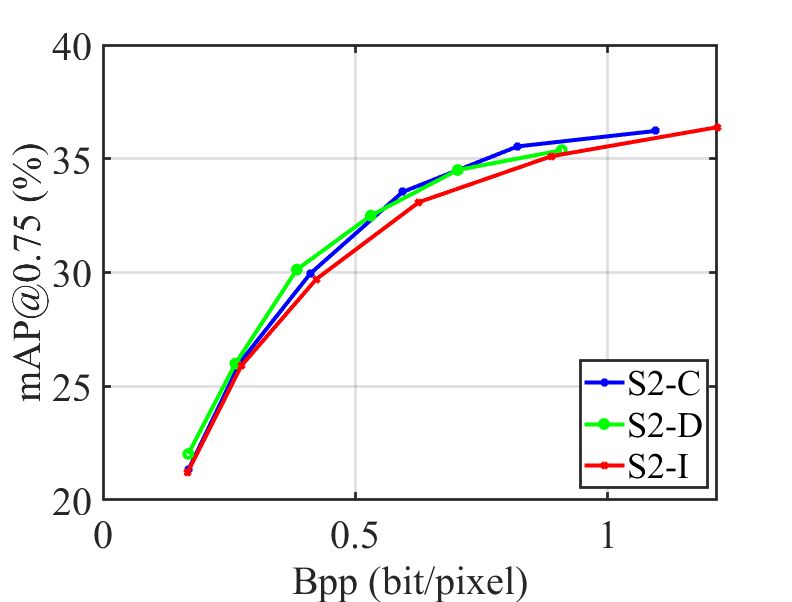}}
%  \vspace{1.5cm}
  \centerline{(c)}\medskip
\end{minipage}
\vspace{-4mm}
\caption{The performance comparison among the three distributions for stage 2 in terms of (a) rate-mAP@0.50:0.95; (b) rate-mAP@0.50; (c) rate-mAP@0.75.}
\vspace{-2mm}
\label{stage2}
\end{figure*}

\section{Discussions}
The proposed scheme has close connections to a number of image compression, feature compression and analysis methods.
\begin{itemize}
    \item \textbf{Connections to image compression.} The state-of-the-art image compression schemes remove the redundancy in terms of spatial, statistics and perception. The proposed scheme still inherits the image compression pipeline within CTA framework, and attempts to further remove the redundancies from the perspective of machine vision. In this vein, it is interesting to find that there is still large room to improve the coding performance in terms of RAO. 
    %The proposed scheme still inherits the image compression pipeline within the CTA framework, and attempts to further remove the redundancies from the perspective of machine vision.
    \item \textbf{Connections to feature compression.} Though the  feature compression could lead to more compact representation, it does not guarantee the reconstruction of the texture and lacks the generalization capability to a certain range of analysis tasks. 
    The proposed scheme is optimized for machine vision, but still preserves the capability of signal level reconstruction. This brings the advantages of enhanced generalization capability to unseen tasks and human-level monitoring. 
    %state-of-the-art image compression schemes remove the redundancy in terms of spatial, statistics and perception. The proposed scheme still inherits the image compression pipeline, and attempts to further remove the redundancies from the perspective of machine vision. In this vein, it is interesting to find that there is still large room to improve the coding performance in terms of RAO. 
    \item \textbf{Connections to analysis methods.} The proposed scheme treats object detection as the analysis task, due to the fact that object detection is the foundation of numerous machine vision applications. It is envisioned that in the future with the increase in the scale of deep learning models, most machine vision tasks are expected to be unified with one architecture. This could bring more feasibilities to the design of the coding scheme towards machine vision, in particular from the standardization perspective.  
    
    %the  feature compression could be more compact, it does not guarantee the reconstruction of the texture and lacks the generalization capability to a certain range of analysis tasks. 
    %The proposed scheme is optimized with the machine vision, but still preserves the capability of signal level reconstruction.  
\end{itemize}

\setlength{\tabcolsep}{8mm}{
\begin{table*}[tb]
\caption{The number of parameters {and the encoding time} for the encoder with various channel distributions. Herein, the parameter number is denoted as ``$\#$ of Param.'' in the table. {For every channel distribution, the encoding time is averaged with various quality levels. The proposed inverted bottleneck encoder with channel distribution S2-D achieves 48.23\% and 35.35\% reduction in terms of parameter number and encoding time respectively, compared with the constant channel distribution S1-C.}}
\vspace{-2mm}
\begin{tabular}{|c|c|c|c|c|c|}
\hline
& S1-C    & S1-D    & S2-C(S1-I) & S2-D    & S2-I    \\ \hline
$\#$ of Param.   & 2,891,136 & 4,428,928 & 1,803,904    & \textbf{1,496,640} & 2,111,168 \\ \hline
Encoding time (second) & 2.778 & 3.694 & 1.986 & \textbf{1.796} & 2.288 \\ \hline
\end{tabular}
%\vspace{-4mm}
\label{param}
\end{table*}}

\section{Experiments}

\begin{figure*}[t]
\begin{minipage}[b]{0.3\linewidth}
  \centering
  \centerline{\includegraphics[width=6.2cm]{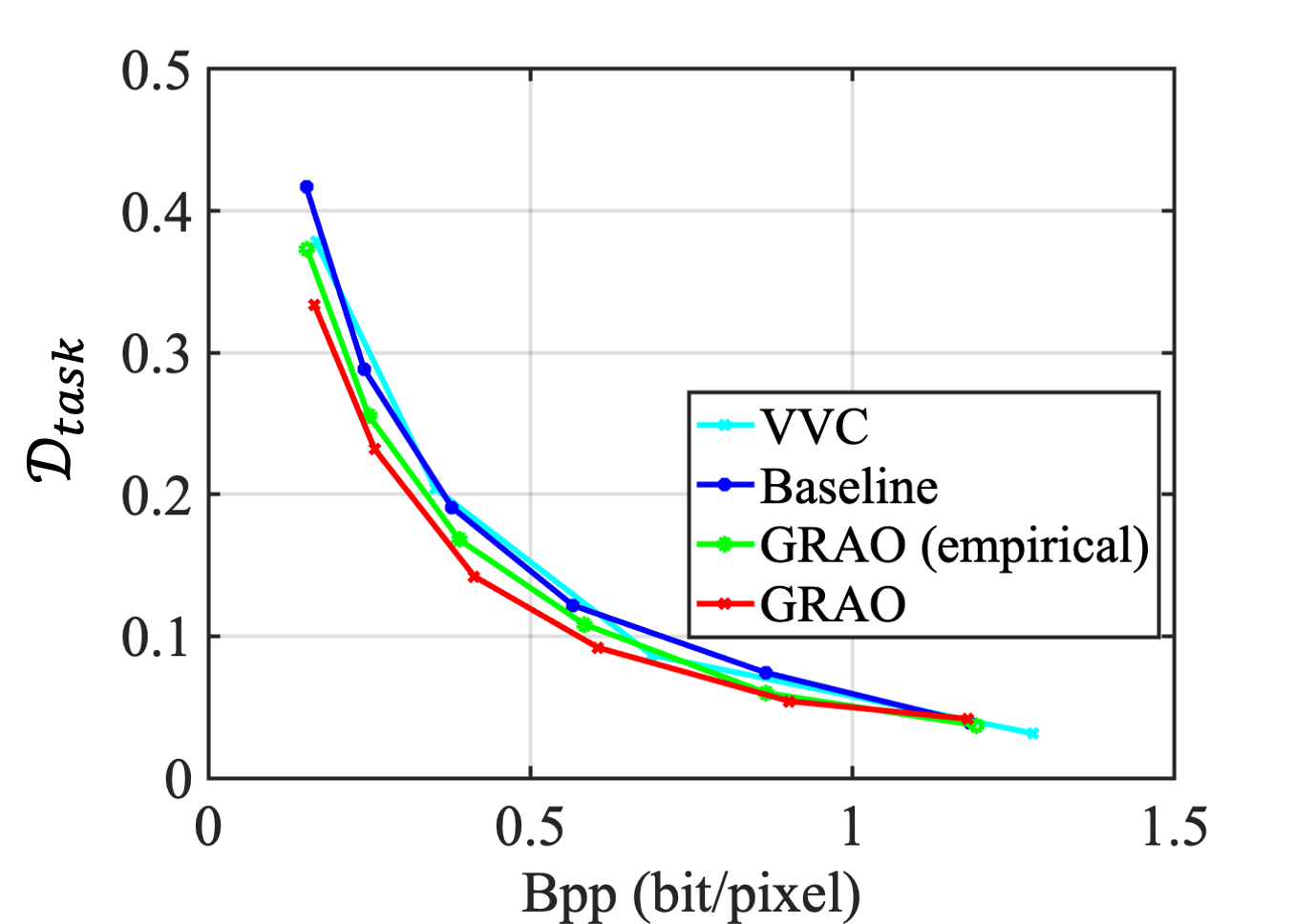}}
%  \vspace{1.5cm}
  \centerline{(a)}\medskip
\end{minipage}
\hfill
\begin{minipage}[b]{0.3\linewidth}
  \centering
  \centerline{\includegraphics[width=6.0cm]{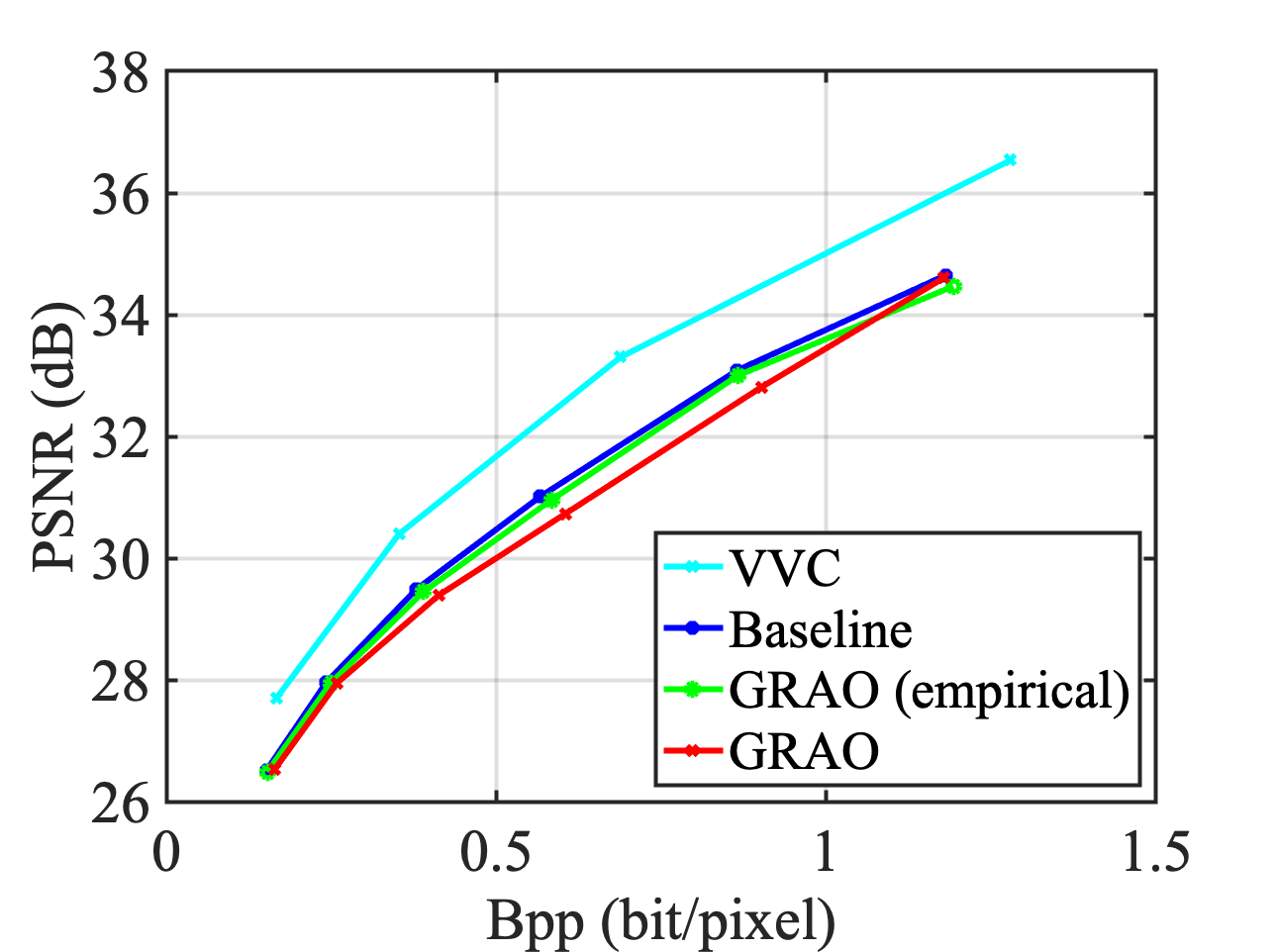}}
%  \vspace{1.5cm}
  \centerline{(b)}\medskip
\end{minipage}
\hfill
\begin{minipage}[b]{0.3\linewidth}
  \centering
  \centerline{\includegraphics[width=6.0cm]{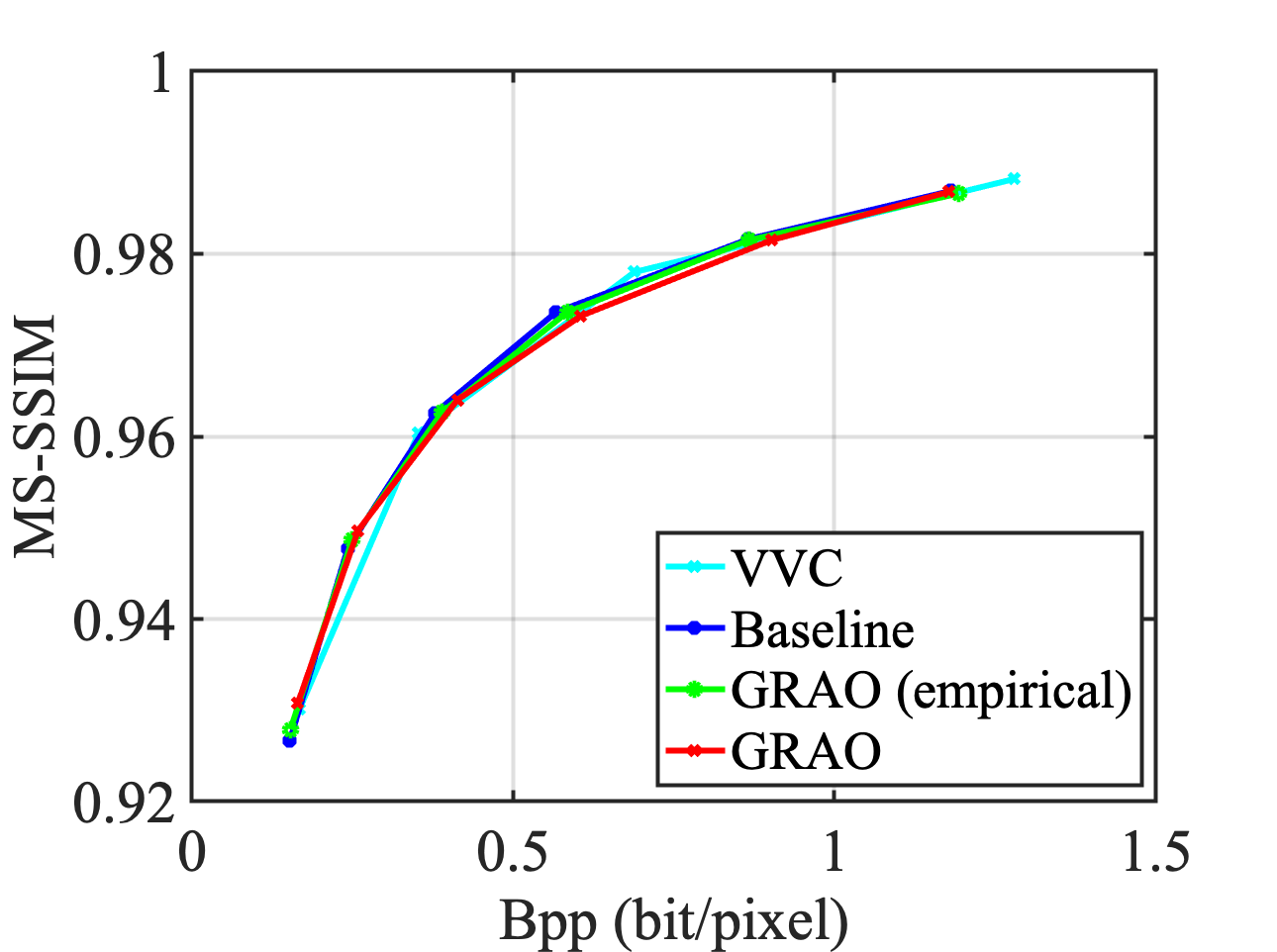}}
%  \vspace{1.5cm}
  \centerline{(c)}\medskip
\end{minipage}
\vspace{-4mm}
\caption{The rate-distortion performance comparison among proposed GRAO, GRAO (empirical), Baseline, and VVC in terms of (a) $\mathcal{D}_{task}$; (b) PSNR; (c) MS-SSIM. }
\vspace{-2mm}
\label{grao}
\end{figure*}

\begin{figure*}[t]
\begin{minipage}[b]{0.3\linewidth}
  \centering
  \centerline{\includegraphics[width=6.0cm]{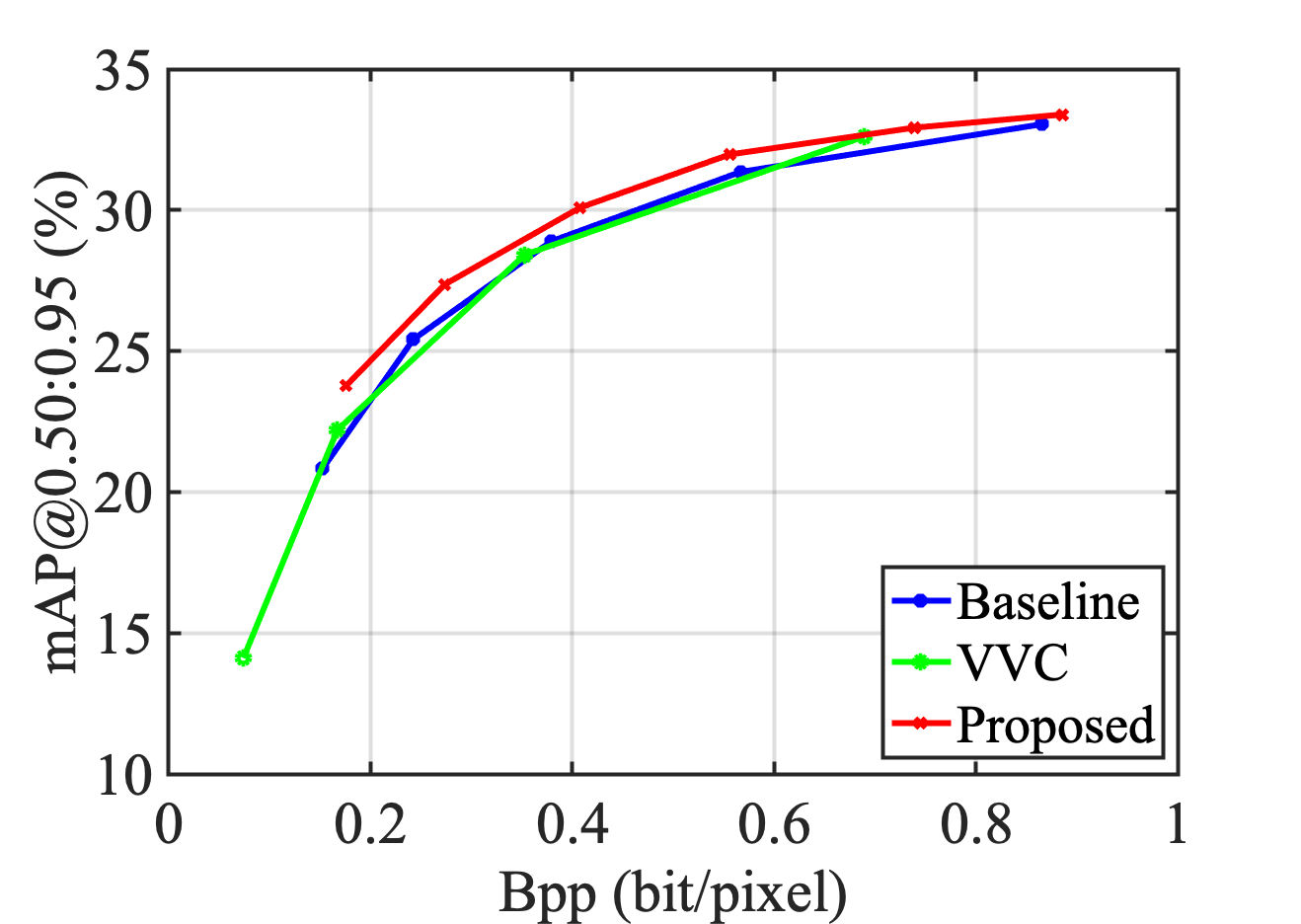}}
%  \vspace{1.5cm}
  \centerline{(a)}\medskip
\end{minipage}
\hfill
\begin{minipage}[b]{0.3\linewidth}
  \centering
  \centerline{\includegraphics[width=6.0cm]{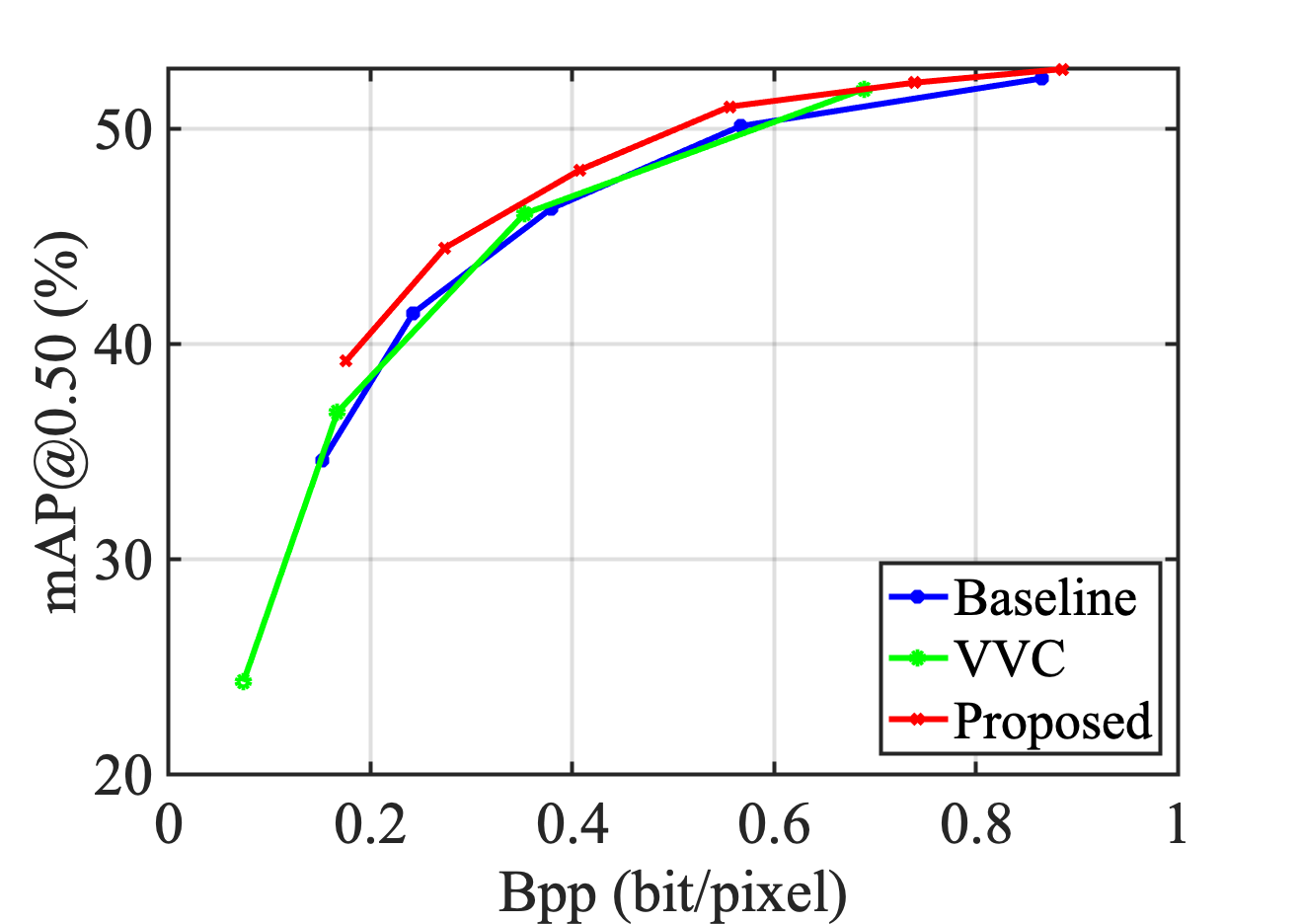}}
%  \vspace{1.5cm}
  \centerline{(b)}\medskip
\end{minipage}
\hfill
\begin{minipage}[b]{0.3\linewidth}
  \centering
  \centerline{\includegraphics[width=6.0cm]{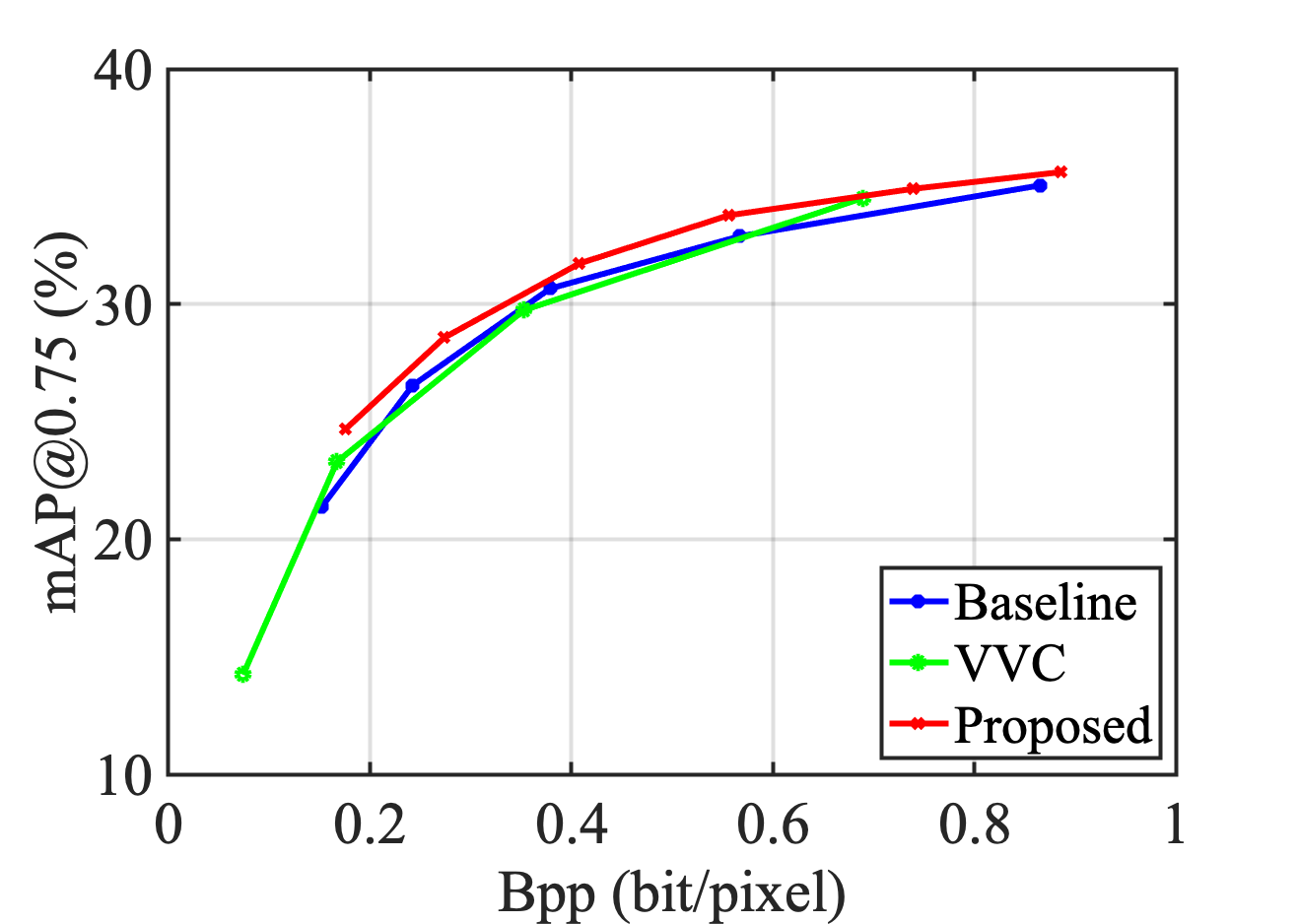}}
%  \vspace{1.5cm}
  \centerline{(c)}\medskip
\end{minipage}
\vspace{-4mm}
\caption{The performance comparison of the proposed algorithm with Baseline and VVC in terms of (a) rate-mAP@0.50:0.95; (b) rate-mAP@0.50; (c) rate-mAP@0.75.}
\vspace{-2mm}
\label{all}
\end{figure*}

To validate the efficiency of the proposed scheme, we evaluate the rate-accuracy performance of the proposed algorithm under various compression quality levels, compared with the state-of-the-art codecs. The experimental setup of the proposed algorithm is first introduced. Subsequently, the experimental results of the proposed inverted bottleneck structure are presented, and the computational complexity is also 
analyzed. Moreover, the performance of the proposed generalized RAO is investigated in terms of machine vision and the generalization capability towards other tasks is validated from the perspective of the signal-level reconstruction performance. Finally, the image compression performance towards object detection with the combination of proposed inverted bottleneck encoder and generalized rate-accuracy optimization is presented, comparing with state-of-the-art image coding schemes including VVC and end-to-end image compression. 

\subsection{Experimental Setup}
The end-to-end image compression towards machine vision is implemented using PyTorch \cite{paszke2019pytorch}. For the image compression model, the learning rate of the en/decoder and the entropy model are set to $0.0001$ and $0.001$ respectively. The optimization method is Adaptive Moment Estimation (Adam) \cite{kingma2014adam}. 
The training and testing data are the training and validation part of COCO2017 dataset \cite{lin2014microsoft} respectively. 
%The training data is COCO2017 dataset training fold and the object detection performance is evaluated with the corresponding COCO2017 dataset validation fold \cite{lin2014microsoft}. 
In order to achieve stable compression performance, the end-to-end image compression model is first trained under the conventional rate-distortion loss function, where $\mathcal{L}_{task}$ is not involved. The batch size and the patch size are 16 and $256\times 256$ respectively. The model is trained with 70 epochs and the setting of $\lambda_{1}$ under various compression levels is the same as \cite{begaint2020compressai}. 
%\sr{More specifically, modules in image compression is identical with \cite{minnen2018joint}, except for encoder $E$.} 
Regarding the optimization of image compression with the proposed joint loss function, the parameter in the codec is initialized with the pretrained model with the same $\lambda_{1}$ value. The parameters of Faster-RCNN are initialized and fixed with the released model in \cite{wu2019detectron2}. The patch size is enlarged to $512\times 512$ to contain the objects with multiple scales. The batch size is set to 8 with the training epoch 1. Moreover, the initial $\lambda_{2}$ and interval $d$ are empirically set to 40 and 32 respectively. The shrinking factor $w$ and the maximal iteration number $N$ are 2 and 4, respectively. We evaluate the compression performance of VVC using the open source implementation VVenC \cite{fraunhofer2020vvenc} with YUV444 format under slower preset.   
\subsection{Performance of Inverted Bottleneck Encoder}

We divide the evaluation into two stages, %including 
%evaluate the performance of inverted bottleneck encoder in terms of object detection in this section. The evaluation is divided into two stages and 
where the first stage focuses on the effectiveness of the monotonically increasing channel numbers comparing with other solutions (monotonically decreasing and constant). Grounded on this design, the second stage targets to  investigate the advantage by decreasing the channel number, which is further compared to the solution with increasing and constant channel numbers. More specifically, the channel distribution under these two stages is listed in Table \ref{invbt}. 

\begin{figure*}[t]
\centerline{\includegraphics[width=6.8in]{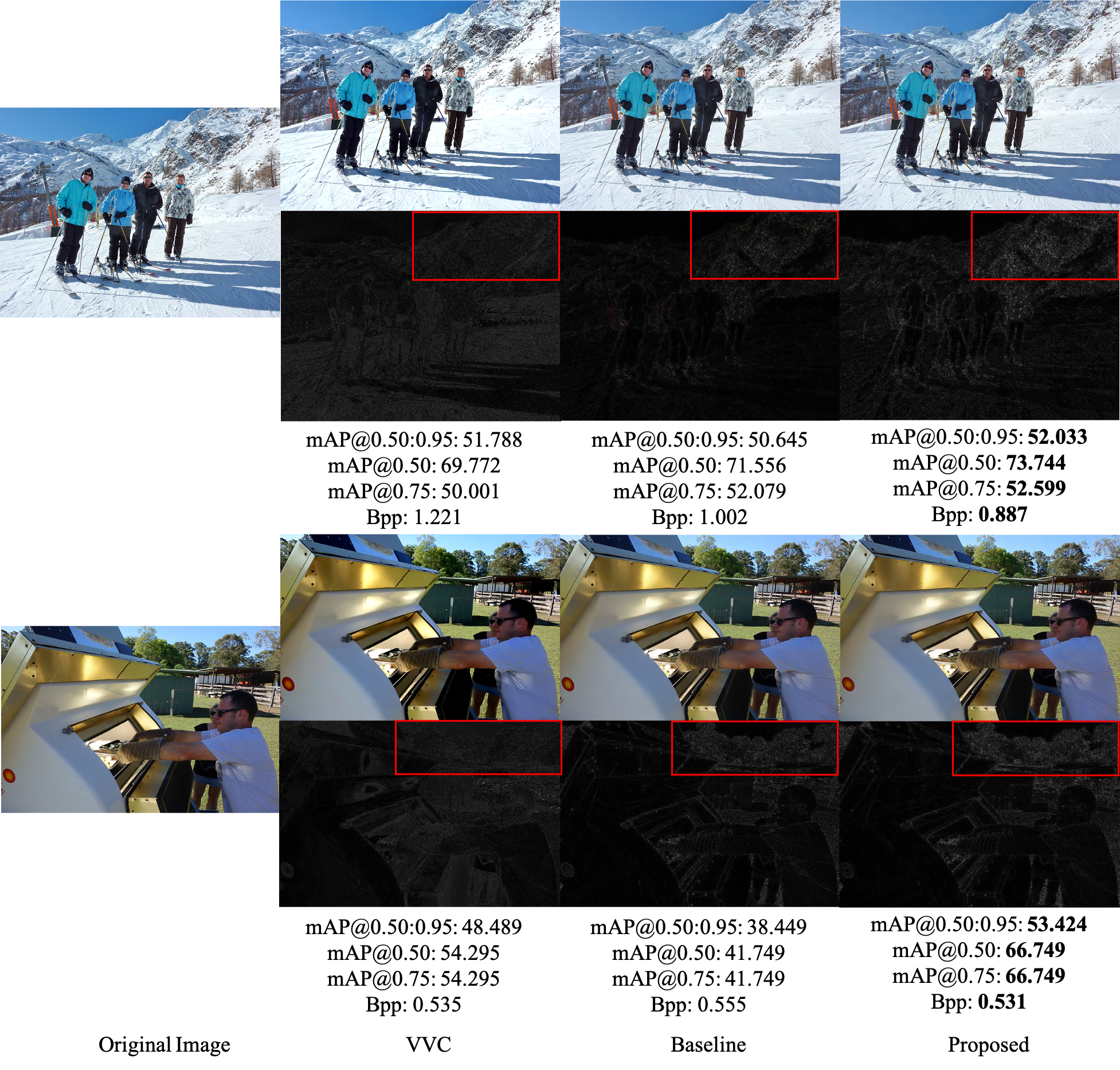}}
\vspace{-2mm}
\caption{{Visualization of the reconstructed images of the proposed framework, compared with the VVC and baseline model. For every compression codec, the MAD between the reconstructed and the original images are visualized with min-max normalization, accompanying with the bitrate and the analysis accuracy. The red rectangles indicate the regions that are compressed aggressively, as they are not particularly important in the analysis task.}}
\label{perceptual}
\vspace{-4mm}
\end{figure*}

\begin{figure*}[th]
\begin{minipage}[b]{0.3\linewidth}
  \centering
  \centerline{\includegraphics[width=6.0cm]{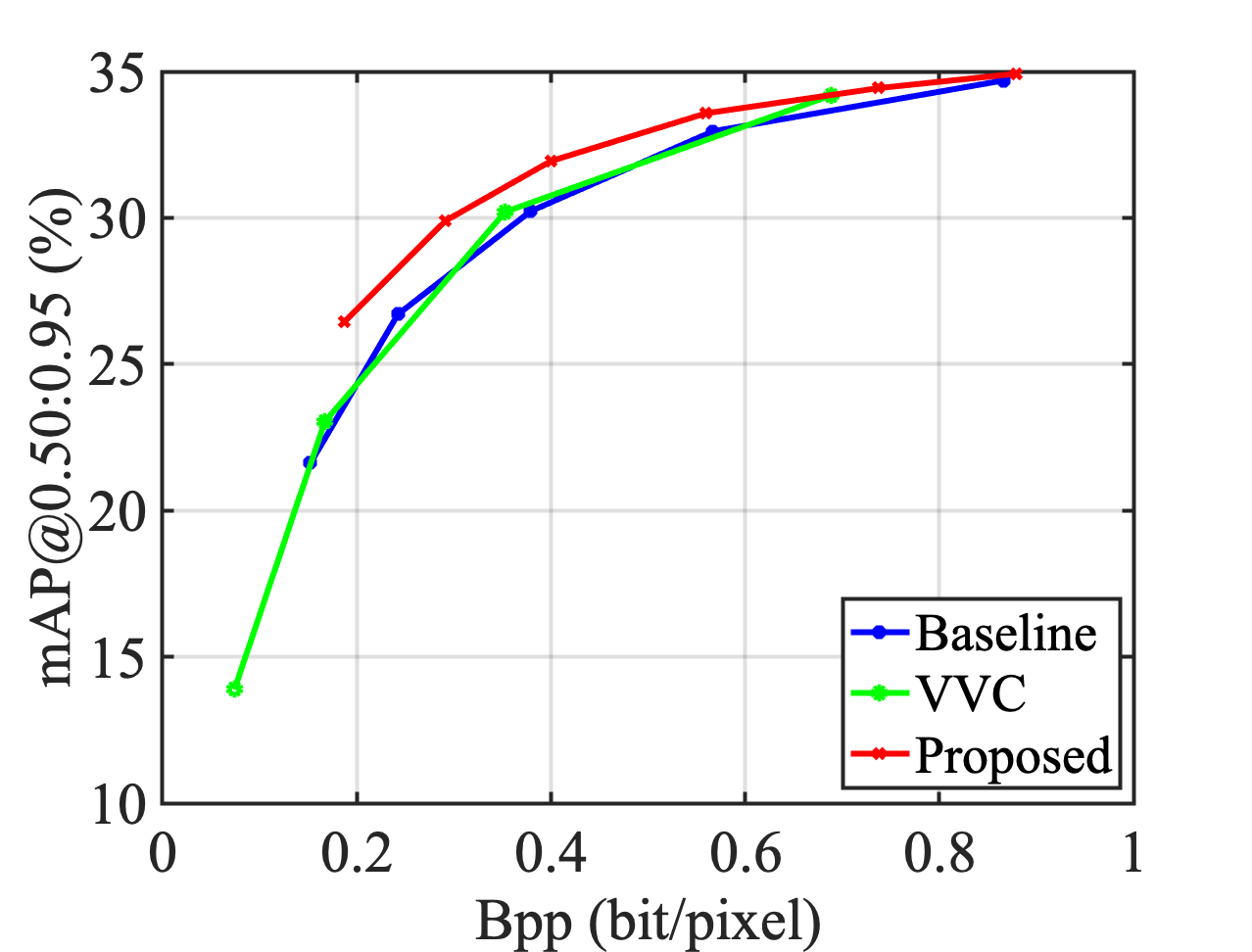}}
%  \vspace{1.5cm}
  \centerline{(a)}\medskip
\end{minipage}
\hfill
\begin{minipage}[b]{0.3\linewidth}
  \centering
  \centerline{\includegraphics[width=6.0cm]{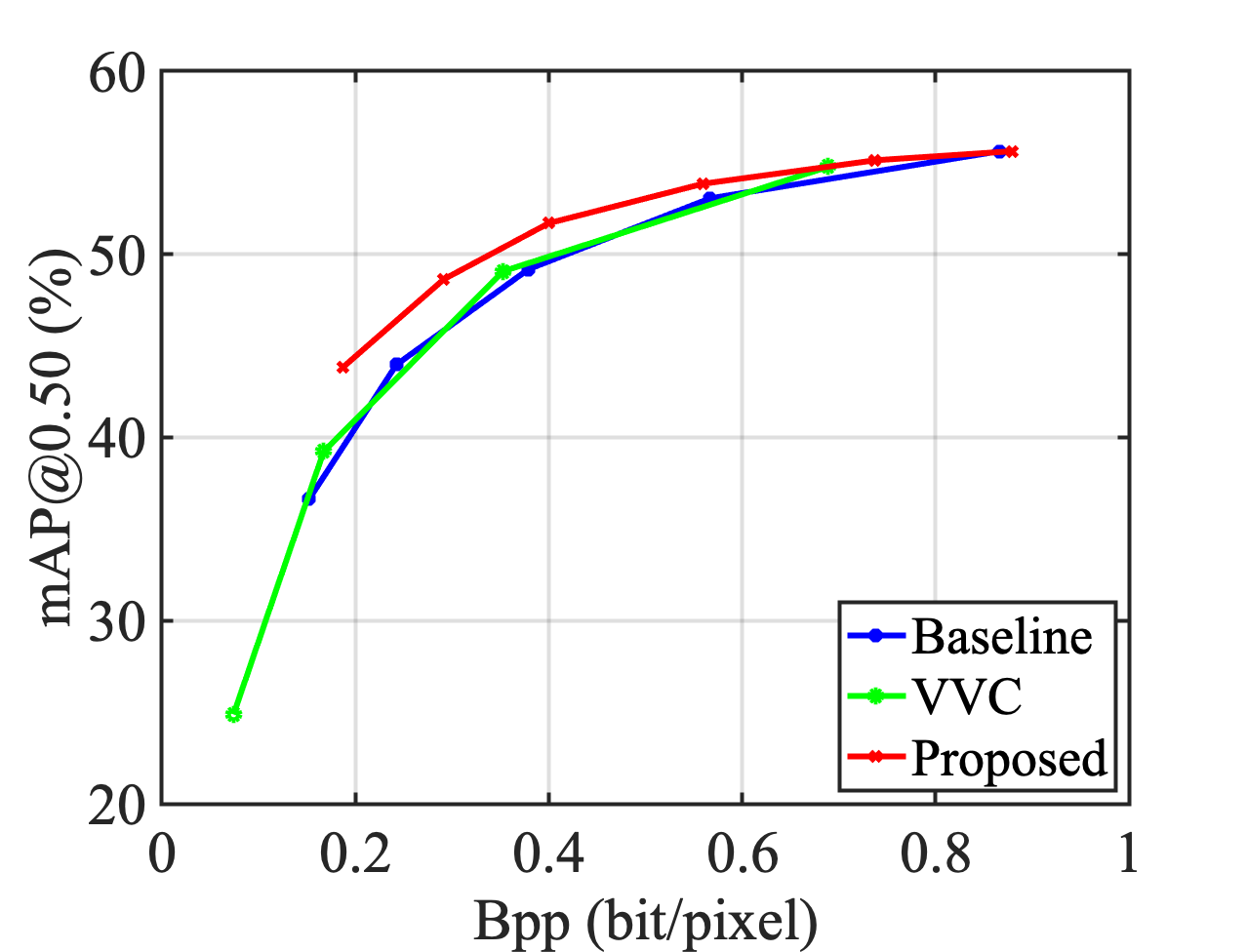}}
%  \vspace{1.5cm}
  \centerline{(b)}\medskip
\end{minipage}
\hfill
\begin{minipage}[b]{0.3\linewidth}
  \centering
  \centerline{\includegraphics[width=6.0cm]{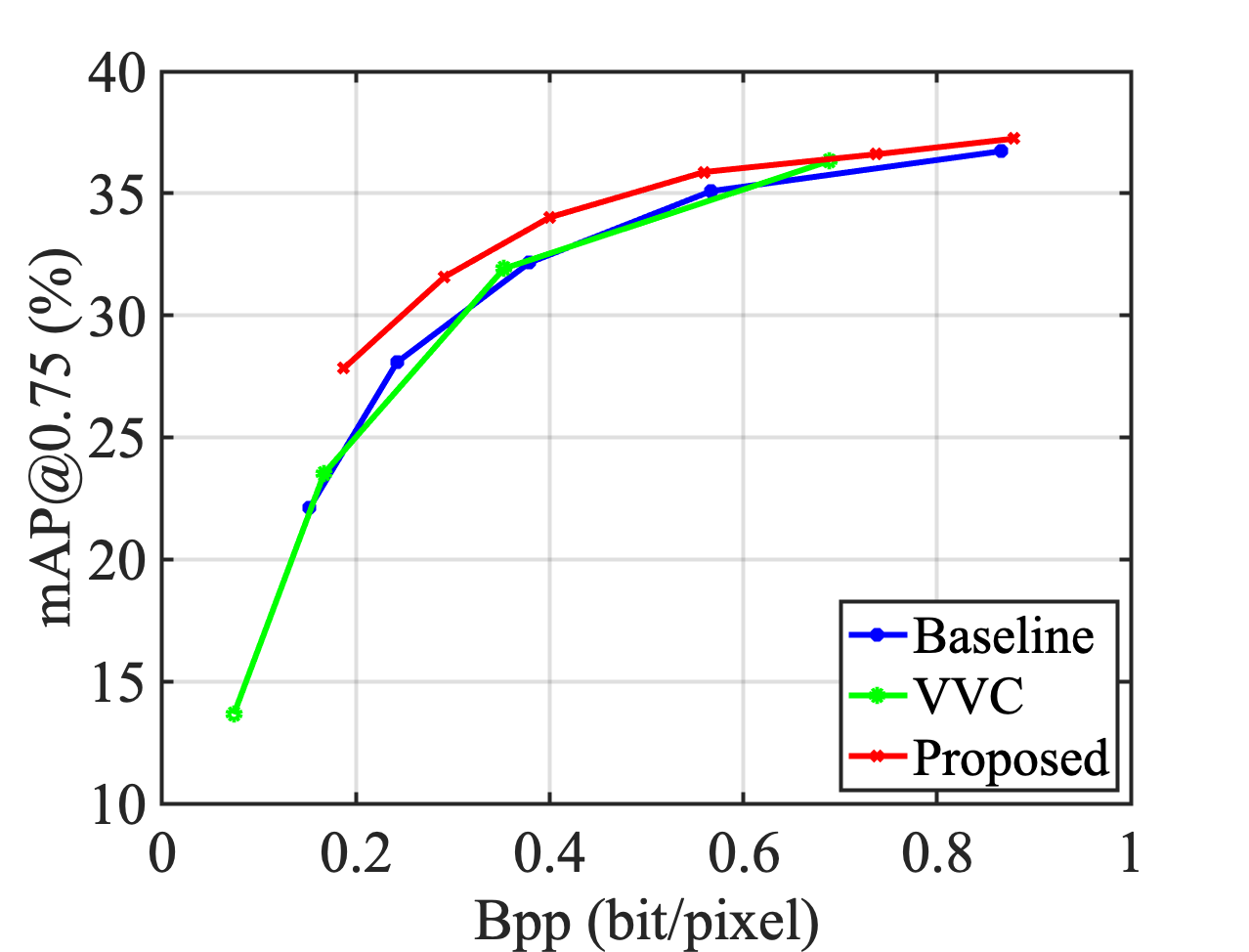}}
%  \vspace{1.5cm}
  \centerline{(c)}\medskip
\end{minipage}
\vspace{-4mm}
\caption{The performance comparison for instance segmentation of the proposed algorithm with Baseline and VVC in terms of (a) rate-mAP@0.50:0.95; (b) rate-mAP@0.50; (c) rate-mAP@0.75.}
\vspace{-2mm}
\label{seg}
\end{figure*}

For the first stage, we evaluate the rate-analysis performance in terms of mAP@0.5:0.95, mAP@0.50 and mAP@0.75, all of which have been widely used in object detection evaluation \cite{ren2015faster, wang2018pelee}. As shown in Fig. \ref{stage1}, the compression model with three distributions (S1-C, S1-D and S1-I) reveal very close performance under various analysis accuracy measures. It is apparent that for the model S1-I which economizes the channels could achieve the comparable performance with lower computational complexity. 
These experimental results provide useful evidence on the effectiveness of the proposed monotonically increasing structure for feature extraction.
Furthermore, based on such design of channel distribution in stage 1, which is fixed as the monotonically increasing distribution, we evaluate the object detection performance under various bitrates of the three distributions in the stage 2. As shown in Fig. \ref{stage2}, comparable performance has also been achieved under the three distributions and the encoder structure with S2-D distribution achieves better performance at low bitrate. As shown in Table \ref{param}, the inverted bottleneck structure with channel distribution S2-D could reduce the computational expense with the fewest number of parameters, {which is also verified with encoding time consumption}. Consequently, by combining the stage 1 and stage 2, the proposed inverted bottleneck structure {with channel distribution S2-D} could compactly represent the images with comparable representation efficiency and lower computational complexity.

\subsection{Performance of Generalized Rate-accuracy Optimization}

In this subsection, we evaluate the performance of proposed  generalized RAO (GRAO) framework. 
%and exploit the relationship between the two weighting factors, $\lambda_{1}$ and $\lambda_{2}$. 
In particular, the performance comparison is achieved in terms of the rate-$\mathcal{D}_{task}$ performance, and the compared methods include the method 
%Regarding the performance improvement of RAO, we compare the , comparing with the performance 
without GRAO and the empirical selected $\lambda_{2}$ value with GRAO, denoted as Baseline and GRAO (empirical) respectively. Specifically, the loss function of Baseline model is $\mathcal{L}=\lambda_{1}\mathcal{L}_{mse}+\mathcal{R}$ and the loss function of GRAO-based models follows the formulation of Equ. \ref{equ1}.
As shown in Fig. \ref{grao}.(a), there is an obvious performance improvement of the proposed GRAO framework and the proposed iterative GRAO can further improve the compression performance comparing with GRAO (empirical). In particular, the performance gain is more obvious for low bit rate scenarios due to the fact that the optimization plays a more important role in low bit rate coding scenarios. In contrast, the performance gain saturates at high bit rates as the degradation of image compression for analysis tasks is not obvious in high bit rate scenarios.
%the basis of the selected $\lambda_{2}$ values after the iterations,
%The relationship between $R$ and $\lambda_{2}$ is shown in  with the exponential function, shown in Fig. \ref{fit}, which is well-fitted and the R2-score is 0.9891. 
The signal-level reconstruction performance of the proposed GRAO framework is also shown in Fig. \ref{grao}(b) in terms of rate-PSNR. A signal-level representation performance with acceptable performance degradation compared with Baseline is observed. This further demonstrates the promising generalization capability of the proposed scheme towards other tasks due to the signal-level information representation.

\subsection{The Overall Performance Evaluation}

We combine the proposed inverted bottleneck structure and the off-line searching algorithm together to evaluate the performance improvement in terms of machine vision with various measures. Herein, we directly apply the $\lambda_{2}$ values in Section 6.3, in an effort to study the straightforward combination of these two schemes.
%adopted directly in the optimization of inverted bottleneck encoder with the joint loss function. 
The performance is shown in Fig. \ref{all} and it confirms the proposed scheme could achieve significant performance improvement under various evaluation metrics, especially at low bitrates. Specifically, comparing with VVC, the proposed scheme could achieve 9.06\% bit rate savings in terms of mAP@0.50:0.95 with Bj\o ntegaard-Delta rate \cite{vcmbdrate}. Moreover, compared to the end-to-end compression without any modification, the proposed scheme can also achieve significant coding bits savings, revealing the promise of the proposed scheme in a variety of machine vision applications. {The visualization of the reconstructed images is also provided in Fig. \ref{perceptual}. The MAD between the original and reconstructed images of various compression codecs are 
shown. It can be observed that the regions that are not particularly important in the analysis task could be substantially compressed in the proposed scheme, revealing the design philosophy of the proposed coding technique.}

\subsection{The Performance of Generalization Capability}
The generalization capability is further investigated in this subsection. In principle, the loss function of object detection in the optimization scheme tends to preserve the semantic information in the compression, potentially leading to better performance in other analysis tasks. More specifically, we evaluate the instance segmentation performance on the reconstructed images of the proposed scheme comparing with VVC and typical end-to-end image compression \cite{minnen2018joint} under pretrained instance segmentation Mask-RCNN \cite{wu2019detectron2}. As shown in Fig. \ref{seg}, there is an obvious performance improvement for instance segmentation, although the image compression model is designed for object detection task. This also provides more evidence regarding the generalization capability 
of the proposed scheme.

\section{Conclusion}
We propose an end-to-end compression scheme tailored for machine vision, based upon the inverted bottleneck encoding architecture 
%the specifically designed encoding architecture 
and iterative RAO scheme. The novelty of the proposed scheme lies in the new coding network design and the distortion modeling for RAO, which not only ensures the analysis performance but also maintains the capability of signal reconstruction. The benefits of the proposed scheme in terms of computational complexity, rate-accuracy performance and generalization capability are demonstrated using extensive experiments. 

The proposed compression scheme is extensible. For example, more analysis tasks can be combined into the analysis module, making the scheme more sophisticated and powerful. Moreover, the proposed scheme could also be extended to scalable representation, where the output of the end-to-end codec could directly serve as the input module for analysis as the base layer.
One may also improve the proposed scheme by considering more variants of rate-accuracy models for optimization. Moreover, the extension of this codec toward a more
unified compression scheme, in particular for the early feature extraction stage, is also an interesting research direction yet to be explored.

\bibliographystyle{IEEEtran}
\bibliography{IEEEabrv,open}

\begin{thebibliography}{10}
\providecommand{\url}[1]{#1}
\csname url@samestyle\endcsname
\providecommand{\newblock}{\relax}
\providecommand{\bibinfo}[2]{#2}
\providecommand{\BIBentrySTDinterwordspacing}{\spaceskip=0pt\relax}
\providecommand{\BIBentryALTinterwordstretchfactor}{4}
\providecommand{\BIBentryALTinterwordspacing}{\spaceskip=\fontdimen2\font plus
\BIBentryALTinterwordstretchfactor\fontdimen3\font minus
  \fontdimen4\font\relax}
\providecommand{\BIBforeignlanguage}[2]{{%
\expandafter\ifx\csname l@#1\endcsname\relax
\typeout{** WARNING: IEEEtran.bst: No hyphenation pattern has been}%
\typeout{** loaded for the language `#1'. Using the pattern for}%
\typeout{** the default language instead.}%
\else
\language=\csname l@#1\endcsname
\fi
#2}}
\providecommand{\BIBdecl}{\relax}
\BIBdecl

\bibitem{cisco2019cisco}
U.~Cisco, ``Cisco annual internet report (2017--2022) white paper,'' 2019.

\bibitem{cisco2020cisco}
------, ``Cisco annual internet report (2018--2023) white paper,'' 2020.

\bibitem{redondi2013compress}
A.~Redondi, L.~Baroffio, M.~Cesana, and M.~Tagliasacchi,
  ``Compress-then-analyze vs. analyze-then-compress: Two paradigms for image
  analysis in visual sensor networks,'' in \emph{IEEE International Workshop on
  Multimedia Signal Processing}.\hskip 1em plus 0.5em minus 0.4em\relax IEEE,
  2013, pp. 278--282.

\bibitem{duan2015overview}
L.-Y. Duan, V.~Chandrasekhar, J.~Chen, J.~Lin, Z.~Wang, T.~Huang, B.~Girod, and
  W.~Gao, ``Overview of the {MPEG}-{CDVS} standard,'' \emph{IEEE Transactions
  on Image Processing}, vol.~25, no.~1, pp. 179--194, 2015.

\bibitem{duan2018compact}
L.-Y. Duan, Y.~Lou, Y.~Bai, T.~Huang, W.~Gao, V.~Chandrasekhar, J.~Lin,
  S.~Wang, and A.~C. Kot, ``Compact descriptors for video analysis: The
  emerging {MPEG} standard,'' \emph{IEEE MultiMedia}, vol.~26, no.~2, pp.
  44--54, 2018.

\bibitem{xia2020emerging}
S.~Xia, K.~Liang, W.~Yang, L.-Y. Duan, and J.~Liu, ``An emerging coding
  paradigm {VCM}: A scalable coding approach beyond feature and signal,''
  \emph{arXiv preprint arXiv:2001.03004}, 2020.

\bibitem{wallace1992jpeg}
G.~K. Wallace, ``The {JPEG} still picture compression standard,'' \emph{IEEE
  Transactions on Consumer Electronics}, vol.~38, no.~1, pp. xviii--xxxiv,
  1992.

\bibitem{rabbani2002jpeg2000}
M.~Rabbani, ``{JPEG}2000: Image compression fundamentals, standards and
  practice,'' \emph{Journal of Electronic Imaging}, vol.~11, no.~2, p. 286,
  2002.

\bibitem{lian2012webp}
L.~Lian and W.~Shilei, ``Webp: A new image compression format based on vp8
  encoding,'' \emph{Microcontrollers \& Embedded Systems}, vol.~3, 2012.

\bibitem{wiegand2003overview}
T.~Wiegand, G.~J. Sullivan, G.~Bjontegaard, and A.~Luthra, ``Overview of the
  {H}. 264/{AVC} video coding standard,'' \emph{IEEE Transactions on Circuits
  and Systems for Video Technology}, vol.~13, no.~7, pp. 560--576, 2003.

\bibitem{sullivan2012overview}
G.~J. Sullivan, J.-R. Ohm, W.-J. Han, and T.~Wiegand, ``Overview of the high
  efficiency video coding ({HEVC}) standard,'' \emph{IEEE Transactions on
  Circuits and Systems for Video Technology}, vol.~22, no.~12, pp. 1649--1668,
  2012.

\bibitem{choi2019design}
Y.-J. Choi, D.-S. Jun, W.-S. Cheong, and B.-G. Kim, ``Design of efficient
  perspective affine motion estimation/compensation for versatile video coding
  ({VVC}) standard,'' \emph{Electronics}, vol.~8, no.~9, p. 993, 2019.

\bibitem{zhang2019recent}
J.~Zhang, C.~Jia, M.~Lei, S.~Wang, S.~Ma, and W.~Gao, ``Recent development of
  {AVS} video coding standard: {AVS}3,'' in \emph{2019 Picture Coding Symposium
  (PCS)}.\hskip 1em plus 0.5em minus 0.4em\relax IEEE, 2019, pp. 1--5.

\bibitem{ramchandran1994rate}
K.~Ramchandran and M.~Vetterli, ``Rate-distortion optimal fast thresholding
  with complete {JPEG}/{MPEG} decoder compatibility,'' \emph{IEEE Transactions
  on image processing}, vol.~3, no.~5, pp. 700--704, 1994.

\bibitem{sullivan1998rate}
G.~J. Sullivan and T.~Wiegand, ``Rate-distortion optimization for video
  compression,'' \emph{IEEE signal processing magazine}, vol.~15, no.~6, pp.
  74--90, 1998.

\bibitem{stankowski2015rate}
J.~Stankowski, C.~Korzeniewski, M.~Doma{\'n}ski, and T.~Grajek,
  ``Rate-distortion optimized quantization in {HEVC}: Performance
  limitations,'' in \emph{2015 Picture coding symposium (PCS)}.\hskip 1em plus
  0.5em minus 0.4em\relax IEEE, 2015, pp. 85--89.

\bibitem{karczewicz2008rate}
M.~Karczewicz, Y.~Ye, and I.~Chong, ``Rate distortion optimized quantization,''
  \emph{ITU-T Q}, vol.~6, 2008.

\bibitem{li2008laplace}
X.~Li, N.~Oertel, A.~Hutter, and A.~Kaup, ``Laplace distribution based
  lagrangian rate distortion optimization for hybrid video coding,'' \emph{IEEE
  Transactions on Circuits and Systems for Video Technology}, vol.~19, no.~2,
  pp. 193--205, 2008.

\bibitem{wang2011ssim}
S.~Wang, A.~Rehman, Z.~Wang, S.~Ma, and W.~Gao, ``{SSIM}-motivated
  rate-distortion optimization for video coding,'' \emph{IEEE Transactions on
  Circuits and Systems for Video Technology}, vol.~22, no.~4, pp. 516--529,
  2011.

\bibitem{toderici2015variable}
G.~Toderici, S.~M. O'Malley, S.~J. Hwang, D.~Vincent, D.~Minnen, S.~Baluja,
  M.~Covell, and R.~Sukthankar, ``Variable rate image compression with
  recurrent neural networks,'' \emph{ICLR}, 2016.

\bibitem{liu2018cnn}
D.~Liu, H.~Ma, Z.~Xiong, and F.~Wu, ``{CNN}-based {DCT}-like transform for
  image compression,'' in \emph{International Conference on Multimedia
  Modeling}.\hskip 1em plus 0.5em minus 0.4em\relax Springer, 2018, pp. 61--72.

\bibitem{balle2015density}
J.~Ball{\'e}, V.~Laparra, and E.~P. Simoncelli, ``Density modeling of images
  using a generalized normalization transformation,'' \emph{arXiv preprint
  arXiv:1511.06281}, 2015.

\bibitem{balle2016end}
J.~Ball{\'e}, V.~Laparra, and E.~Simoncelli, ``End-to-end optimized image
  compression,'' in \emph{International Conference on Learning
  Representations}, 2017.

\bibitem{balle2018variational}
J.~Ball{\'e}, D.~Minnen, S.~Singh, S.~J. Hwang, and N.~Johnston, ``Variational
  image compression with a scale hyperprior,'' in \emph{International
  Conference on Learning Representations}, 2018.

\bibitem{minnen2018joint}
D.~Minnen, J.~Ball{\'e}, and G.~Toderici, ``Joint autoregressive and
  hierarchical priors for learned image compression,'' \emph{arXiv preprint
  arXiv:1809.02736}, 2018.

\bibitem{krizhevsky2012imagenet}
A.~Krizhevsky, I.~Sutskever, and G.~E. Hinton, ``Imagenet classification with
  deep convolutional neural networks,'' in \emph{Advances in Neural Information
  Processing Systems}, 2012, pp. 1097--1105.

\bibitem{simonyan2014very}
K.~Simonyan and A.~Zisserman, ``Very deep convolutional networks for
  large-scale image recognition,'' in \emph{International Conference on
  Learning Representations}, 2015.

\bibitem{szegedy2015going}
C.~Szegedy, W.~Liu, Y.~Jia, P.~Sermanet, S.~Reed, D.~Anguelov, D.~Erhan,
  V.~Vanhoucke, and A.~Rabinovich, ``Going deeper with convolutions,'' in
  \emph{Proceedings of the IEEE conference on computer vision and pattern
  recognition}, 2015, pp. 1--9.

\bibitem{he2016deep}
K.~He, X.~Zhang, S.~Ren, and J.~Sun, ``Deep residual learning for image
  recognition,'' in \emph{Proceedings of the IEEE conference on computer vision
  and pattern recognition}, 2016, pp. 770--778.

\bibitem{ding2020joint}
L.~Ding, Y.~Tian, H.~Fan, C.~Chen, and T.~Huang, ``Joint coding of local and
  global deep features in videos for visual search,'' \emph{IEEE Transactions
  on Image Processing}, vol.~29, pp. 3734--3749, 2020.

\bibitem{wang2020end}
S.~Wang, W.~Yang, and S.~Wang, ``End-to-end facial deep learning feature
  compression with teacher-student enhancement,'' \emph{arXiv preprint
  arXiv:2002.03627}, 2020.

\bibitem{chen2020toward}
Z.~Chen, K.~Fan, S.~Wang, L.~Duan, W.~Lin, and A.~C. Kot, ``Toward intelligent
  sensing: Intermediate deep feature compression,'' \emph{IEEE Transactions on
  Image Processing}, vol.~29, pp. 2230--2243, 2020.

\bibitem{torfason2018towards}
R.~Torfason, F.~Mentzer, E.~Agustsson, M.~Tschannen, R.~Timofte, and
  L.~Van~Gool, ``Towards image understanding from deep compression without
  decoding,'' \emph{arXiv preprint arXiv:1803.06131}, 2018.

\bibitem{duan2020video}
L.-Y. Duan, J.~Liu, W.~Yang, T.~Huang, and W.~Gao, ``Video coding for machines:
  A paradigm of collaborative compression and intelligent analytics,''
  \emph{arXiv preprint arXiv:2001.03569}, 2020.

\bibitem{vcmcfe}
Y.~Z. M.~Rafie and S.~Liu, ``Draft of call for evidence for video coding for
  machines,'' \emph{MPEG doc. m56229 and ISO/IEC JTC 1/SC 29/WG 2}, 2021.

\bibitem{vcmuse}
------, ``Use cases and requirements for video coding for machines,''
  \emph{MPEG doc. m56227 and ISO/IEC JTC 1/SC 29/WG 2}, 2021.

\bibitem{ke2007event}
Y.~Ke, R.~Sukthankar, and M.~Hebert, ``Event detection in crowded videos,'' in
  \emph{2007 IEEE 11th International Conference on Computer Vision}.\hskip 1em
  plus 0.5em minus 0.4em\relax IEEE, 2007, pp. 1--8.

\bibitem{basharat2008learning}
A.~Basharat, A.~Gritai, and M.~Shah, ``Learning object motion patterns for
  anomaly detection and improved object detection,'' in \emph{2008 IEEE
  Conference on Computer Vision and Pattern Recognition}.\hskip 1em plus 0.5em
  minus 0.4em\relax IEEE, 2008, pp. 1--8.

\bibitem{balaji2017survey}
S.~Balaji and S.~Karthikeyan, ``A survey on moving object tracking using image
  processing,'' in \emph{2017 11th international conference on intelligent
  systems and control (ISCO)}.\hskip 1em plus 0.5em minus 0.4em\relax IEEE,
  2017, pp. 469--474.

\bibitem{ren2015faster}
S.~Ren, K.~He, R.~Girshick, and J.~Sun, ``Faster {r}-{cnn}: Towards real-time
  object detection with region proposal networks,'' \emph{arXiv preprint
  arXiv:1506.01497}, 2015.

\bibitem{zeiler2014visualizing}
M.~D. Zeiler and R.~Fergus, ``Visualizing and understanding convolutional
  networks,'' in \emph{European conference on computer vision}.\hskip 1em plus
  0.5em minus 0.4em\relax Springer, 2014, pp. 818--833.

\bibitem{li2018joint}
Y.~Li, C.~Jia, S.~Wang, X.~Zhang, S.~Wang, S.~Ma, and W.~Gao, ``Joint
  rate-distortion optimization for simultaneous texture and deep feature
  compression of facial images,'' in \emph{2018 IEEE Fourth International
  Conference on Multimedia Big Data (BigMM)}.\hskip 1em plus 0.5em minus
  0.4em\relax IEEE, 2018, pp. 1--5.

\bibitem{zhang2016joint}
X.~Zhang, S.~Ma, S.~Wang, X.~Zhang, H.~Sun, and W.~Gao, ``A joint compression
  scheme of video feature descriptors and visual content,'' \emph{IEEE
  Transactions on Image Processing}, vol.~26, no.~2, pp. 633--647, 2016.

\bibitem{ma2018joint}
S.~Ma, X.~Zhang, S.~Wang, X.~Zhang, C.~Jia, and S.~Wang, ``Joint feature and
  texture coding: Toward smart video representation via front-end
  intelligence,'' \emph{IEEE Transactions on Circuits and Systems for Video
  Technology}, vol.~29, no.~10, pp. 3095--3105, 2018.

\bibitem{ding2017rate}
L.~Ding, Y.~Tian, H.~Fan, Y.~Wang, and T.~Huang, ``Rate-performance-loss
  optimization for inter-frame deep feature coding from videos,'' \emph{IEEE
  Transactions on Image Processing}, vol.~26, no.~12, pp. 5743--5757, 2017.

\bibitem{begaint2020compressai}
J.~B{\'e}gaint, F.~Racap{\'e}, S.~Feltman, and A.~Pushparaja, ``Compressai: a
  pytorch library and evaluation platform for end-to-end compression
  research,'' \emph{arXiv preprint arXiv:2011.03029}, 2020.

\bibitem{vcmtest}
Y.~Z. M.~Rafie and S.~Liu, ``Draft of evaluation framework for video coding for
  machines,'' \emph{MPEG doc. m56228 and ISO/IEC JTC 1/SC 29/WG 2}, 2021.

\bibitem{paszke2019pytorch}
A.~Paszke, S.~Gross, F.~Massa, A.~Lerer, J.~Bradbury, G.~Chanan, T.~Killeen,
  Z.~Lin, N.~Gimelshein, L.~Antiga \emph{et~al.}, ``Pytorch: An imperative
  style, high-performance deep learning library,'' \emph{arXiv preprint
  arXiv:1912.01703}, 2019.

\bibitem{kingma2014adam}
D.~P. Kingma and J.~Ba, ``Adam: A method for stochastic optimization,''
  \emph{arXiv preprint arXiv:1412.6980}, 2014.

\bibitem{lin2014microsoft}
T.-Y. Lin, M.~Maire, S.~Belongie, J.~Hays, P.~Perona, D.~Ramanan,
  P.~Doll{\'a}r, and C.~L. Zitnick, ``Microsoft {COCO}: Common objects in
  context,'' in \emph{European conference on computer vision}.\hskip 1em plus
  0.5em minus 0.4em\relax Springer, 2014, pp. 740--755.

\bibitem{wu2019detectron2}
Y.~Wu, A.~Kirillov, F.~Massa, W.-Y. Lo, and R.~Girshick, ``Detectron2,''
  \url{https://github.com/facebookresearch/detectron2}, 2019.

\bibitem{fraunhofer2020vvenc}
H.~Fraunhofer, ``Vvenc software repository,'' 2020.

\bibitem{wang2018pelee}
R.~J. Wang, X.~Li, and C.~X. Ling, ``Pelee: A real-time object detection system
  on mobile devices,'' \emph{arXiv preprint arXiv:1804.06882}, 2018.

\bibitem{vcmbdrate}
L.~L. Christopher~Hollmann, ``On the evaluation of {VCM} proposals,''
  \emph{MPEG doc. m55854 and ISO/IEC JTC 1/SC 29/WG 2}, 2021.

\end{thebibliography}


\begin{thebibliography}{00}
\bibitem{ref1} G. O. Young, ``Synthetic structure of industrial plastics (Book style with paper title and editor),'' in \emph{Plastics}, 2nd ed. vol. 3, J. Peters, Ed. New York: McGraw-Hill, 1964, pp. 15--64.
\bibitem{ref2} W.-K. Chen, \emph{Linear Networks and Systems} (Book style)$. $Belmont, CA: Wadsworth, 1993, pp. 123--135.
\bibitem{ref3} H. Poor, \emph{An Introduction to Signal Detection and Estimation}. New York: Springer-Verlag, 1985, ch. 4.
\bibitem{ref30} R. J. Vidmar. (1992, August). On the use of atmospheric plasmas as electromagnetic reflectors. \emph{IEEE Trans. Plasma Sci.} [Online]. \emph{21(3).} pp. 876--880. Available: http://www.halcyon.com/pub/journals/21ps03-vidmar
\end{thebibliography}

\end{document}